%% file: main.tex
\documentclass[letterpaper, 10 pt, journal, twoside]{IEEEtran} 
\newcommand{\FIGPATH}{./fig/high_res}

\renewcommand{\baselinestretch}{1.00} 
\newcommand{\eqsize}[0]{\small}

\usepackage[T1]{fontenc}

\usepackage{amsmath}
\usepackage{amssymb}
\usepackage{amsthm}
\usepackage{mathtools}

\usepackage{paralist}

\usepackage{verbatim}

\usepackage{color}
\usepackage{graphicx}
\usepackage{adjustbox}

\usepackage{mathptmx}
\usepackage{times}

\usepackage{fancyhdr}

\usepackage{multirow}
\usepackage{makecell}

\usepackage[abs]{overpic}

\newcommand{\argmin}[0]{\operatorname*{arg\,min}}
\newcommand{\argmax}[0]{\operatorname*{arg\,max}}
\newcommand{\minimize}[0]{\operatorname*{minimize}}

\usepackage{algorithm}
\usepackage[noend]{algpseudocode}

\usepackage[us]{datetime}
\usepackage{caption}
\usepackage{subfloat}

\newcommand{\set}[1]{\mathcal{#1}}
\newcommand{\vect}[1]{\mathbf{#1}}
\newcommand{\mat}[1]{\mathbf{#1}}
\newcommand*{\tran}{^{\intercal}}

\usepackage{soul}

\usepackage[normalem]{ulem}

\usepackage[usenames,dvipsnames,table]{xcolor}

\usepackage{todonotes}

\usepackage{subcaption}

\usepackage{latexsym}
\usepackage{color}

\usepackage{cite}

\usepackage[inline]{enumitem} 
\setenumerate[0]{label=\roman*.}

\usepackage{siunitx}
\usepackage{ifthen}

\usepackage{booktabs}
\usepackage{tablefootnote}

\usepackage{subfiles}
\usepackage{bm}

\DeclareMathAlphabet{\mathcal}{OMS}{cmsy}{m}{n}
\SetMathAlphabet{\mathcal}{bold}{OMS}{cmsy}{b}{n}

\setlist[itemize]{noitemsep, nosep}

\usepackage{stackengine} 

\theoremstyle{definition}
\newtheorem{definition}{Definition}

\theoremstyle{remark}
\newtheorem{remark}{Remark}

\newcommand{\leftscript}[2]{\prescript{#1\!}{}{#2}}

\newcommand{\sumC}[2]{\sum_{(\mathbf{p}, \mathbf{q}) \in \mathbb{C}_{#2}^{#1}}}

\usepackage{tikz}
\usepackage{pgfplots}
\usepackage{pgfplotstable}

\usetikzlibrary{calc}
\usepackage{tikz-dimline}

\pgfplotsset{compat=newest}

\newcommand{\tikzcircle}[2][red,fill=red]{\tikz[baseline=-0.5ex]\draw[#1,radius=#2] (0,0) circle ;}%

\definecolor{color_green}{HTML}{008000}
\definecolor{color_red}{HTML}{E0115F}
\definecolor{color_blue}{HTML}{4169E1}
\definecolor{color_red}{rgb}{0.7, 0, 0}
\definecolor{color_green}{rgb}{0.0, 0.7, 0.0}
\definecolor{color_orange}{HTML}{FFA500}

\definecolor{color_pink}{HTML}{952B60}
\definecolor{color_yellow}{HTML}{F1B620}
\definecolor{color_violet}{HTML}{8A6FDF}

\usepackage{pifont}
\newcommand{\greencheck}{{\color{color_green}\checkmark}}
\newcommand{\xmark}{{\color{color_red}{\ding{55}}}}

\newcommand\hlr[1]{#1}
\newcommand\hlg[1]{#1}

\usepackage{hyperref} 
\hypersetup{
  colorlinks,
  citecolor=black,
  filecolor=black,
  linkcolor=black,
  urlcolor=color_blue,
  pdfauthor={},
  pdfsubject={},
  pdftitle={}
}

\def\equationautorefname~#1\null{(#1)\null}

\title{%
\Title}

\newcommand*{\compilewithallfigures}{}%

\usepackage[printonlyused]{acronym}
\acrodef{mav}[UAV]{Unmanned Aerial Vehicle}
\acrodef{ugv}[UGV]{Unmanned Ground Vehicle}
\acrodef{gps}[GPS]{Global Positioning System}
\acrodef{gnss}[GNSS]{global navigation satellite system}
\acrodef{mil}[MIL]{mirrorless interchangeable-lens}
\acrodef{sar}[S\&R]{Search and Rescue}
\acrodef{ctu}[CTU]{Czech Technical University in Prague}
\acrodef{pc}[PC]{personal computer}
\acrodef{lidar}[LiDAR]{light detection and ranging}
\acrodef{imu}[IMU]{inertial measurement unit}
\acrodef{rgb}[RGB]{color}
\acrodef{rgbd}[RGBD]{color-depth}
\acrodef{agl}[AGL]{above ground level}
\acrodef{amsl}[AMSL]{above mean sea level}
\acrodef{loam}[LOAM]{LiDAR Odometry and Mapping}
\acrodef{aloam}[A-LOAM]{Advanced implementation of LOAM}
\acrodef{liosam}[LIO-SAM]{LiDAR Inertial Odometry via Smoothing and Mapping}
\acrodef{dof}[DoF]{degrees of freedom}
\acrodef{lkf}[LKF]{linear Kalman filter}
\acrodef{mems}[MEMS]{micro-electromechanical systems}
\acrodef{fov}[FoV]{field of view}
\acrodef{slam}[SLAM]{simultaneous localization and mapping}
\acrodef{vio}[VIO]{visual-inertial odometry}
\acrodef{tio}[TIO]{thermal-inertial odometry}
\acrodef{lio}[LIO]{LiDAR-inertial odometry}
\acrodef{icp}[ICP]{Iterative Closest Point}
\acrodef{gicp}[GICP]{Generalized Iterative Closest Point}
\acrodef{ape}[APE]{Absolute Position Error}
\acrodef{ate}[ATE]{Absolute Trajectory Error}
\acrodef{gmm}[GMM]{Gaussian Mixture Models}
\acrodef{rssi}[RSSI]{Received Signal Strength Indicator}
\acrodef{rti}[RTI]{reflectance transformation imaging}
\acrodef{ptm}[PTM]{polynomial texture map}
\acrodef{uv}[UV]{ultraviolet}
\acrodef{ir}[IR]{infrared}
\acrodef{vis}[VIS]{visible spectrum photography}
\acrodef{vistr}[VISTR]{visible spectrum transmitography}
\acrodef{rak}[RAK]{raking light}
\acrodef{tpl}[TPL]{three point lighting}
\acrodef{vivl}[VIVL]{light-induced luminescence}
\acrodef{uvr}[UVR]{UV reflectography}
\acrodef{uvf}[UVF]{UV fluorescent photography}
\acrodef{uvrfc}[UVRFC]{false-color UV reflectography}
\acrodef{irr}[IRR]{IR reflectography}
\acrodef{irrtr}[IRRTR]{IR transmitography}
\acrodef{irf}[IRF]{IR fluorescent photography}
\acrodef{irrfc}[IRRFC]{false-color IR reflectography}
\acrodef{tsp}[TSP]{Traveling Salesman Problem}
\acrodef{ooi}[OoI]{object of interest}
\acrodef{mpc}[MPC]{model predictive control}
\acrodef{gfh}[GFH]{\textit{gradient flow heuristic}}
\acrodef{pdf}[PDF]{probability density function}
\acrodef{pcl}[PCL]{Point Cloud Library}

\acused{lidar}
\acused{dof}

\newcommand{\Title}{RMS: Redundancy-Minimizing Point Cloud Sampling for Real-Time Pose Estimation} 

\author{Pavel Petracek$^{1}$, Kostas Alexis$^{2}$, and Martin Saska$^1$

\thanks{%
This work was supported
by CTU grant no. SGS23/177/OHK3/3T/13,
by the Czech Science Foundation under research project No. 23-06162M,
by the European Union under the project Robotics and advanced industrial production (reg. no. CZ.02.01.01/00/22\_008/0004590), and
by the Research Council of Norway Award NO-321435.}%
  \thanks{$^{1}$ Authors are with the Department of Cybernetics, Faculty of Electrical Engineering, Czech Technical University, Czech Republic {(corresonding author: \tt\footnotesize \href{mailto:pavel.petracek@fel.cvut.cz}{pavel.petracek@fel.cvut.cz}}).}%
  \thanks{$^{2}$ Author is with the Autonomous Robots Lab, Norwegian University of Science and Technology, O. S. Bragstads Plass 2D, 7034, Trondheim, Norway {(\tt\footnotesize \href{mailto:konstantinos.alexis@ntnu.no}{konstantinos.alexis@ntnu.no}}).}%
}

\begin{document}

\setlength{\abovedisplayskip}{4pt}
\setlength{\belowdisplayskip}{3.5pt}


\markboth{IEEE Robotics and Automation Letters. Preprint Version. Accepted March, 2024}
{Petracek \MakeLowercase{\textit{et al.}}: RMS: Redundancy-Minimizing Point Cloud Sampling for Real-Time Pose Estimation}  

 
\maketitle

\begin{abstract}
  \hlr{%
  The typical point cloud sampling methods used in state estimation for mobile robots preserve a high level of point redundancy.
  This redundancy unnecessarily slows down the estimation pipeline and may cause drift under real-time constraints.
  Such undue latency becomes a bottleneck for resource-constrained robots (especially \acsp{mav}), requiring minimal delay for agile and accurate operation.
  We propose a novel, deterministic, uninformed, and single-parameter point cloud sampling method named RMS that minimizes redundancy within a 3D point cloud.
  In contrast to the state of the art, RMS balances the translation-space observability by leveraging the fact that linear and planar surfaces inherently exhibit high redundancy propagated into iterative estimation pipelines.
  We define the concept of \textit{gradient flow}, quantifying the local surface underlying a point.
  We also show that maximizing the entropy of the \textit{gradient flow} minimizes point redundancy for robot ego-motion estimation.
  We integrate RMS into the \textit{point}-based KISS-ICP and \textit{feature}-based LOAM odometry pipelines and evaluate experimentally on KITTI, Hilti-Oxford, and custom datasets from multirotor \acsp{mav}.
  The experiments demonstrate that RMS outperforms state-of-the-art methods in speed, compression, and accuracy in well-conditioned as well as in geometrically-degenerated settings.}
\end{abstract}



\vspace{-1mm}

\vspace{-4mm}
\section*{Multimedia Materials}
\noindent
The paper is supported by code and multimedia materials available at \href{https://github.com/ctu-mrs/RMS}{github.com/ctu-mrs/RMS}.

\vspace{-3mm}
\section{Introduction}
\label{sec:introduction}



\IEEEPARstart{F}{or} the accurate and real-time ego-motion estimation of a resource-constrained robot, the amount of data provided in a 3D \acs{lidar} point cloud is plentiful.
To achieve convergence under real-time constraints (i.e., number of iterations, comp. time, convergence rate), the point clouds must be reduced.
\hlr{%
  Apart from cardinality reduction, the objectives of such point cloud sampling are twofold --- preserve the quality of the points and be computationally fast.
  While the latter is subject to algorithm efficiency and available computational resources, the former must preserve the overall information available in the point cloud.
  In the task of point cloud matching, the contribution (i.e., information) of a point has been shown to be quantifiable via its point-map correspondence and the shape of the loss function~\cite{zhang2016DegeneracyOptimizationbasedState,jiao2021GreedyBasedFeatureSelection, li2021KFSLIOKeyFeatureSelection, tuna2024XICPLocalizabilityAwareLiDAR}. 
  However, information-aware sampling of an input point cloud without the knowledge of these point-map correspondences (uninformed sampling) is non-causal and remains an ongoing challenge.
}

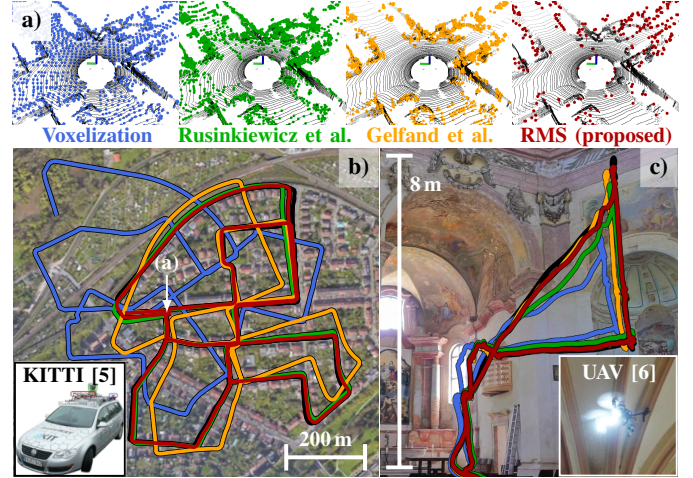
\begin{figure}
  \begin{minipage}[t]{1.0\columnwidth}
    \centering
    \subfloat{\hspace{0mm}\vspace{0mm}\input{./fig/motivation/sampling_frames.tex}}
  \end{minipage}
  \vspace{-5mm}\\
  \begin{minipage}[t]{1.0\columnwidth}
  \centering
    \subfloat{\hspace{0mm}\vspace{0mm}\input{./fig/motivation/labels.tex}}
  \end{minipage}
  \vspace{-5.2mm}\\
  \begin{minipage}[t]{0.548\columnwidth}
  \centering
    \subfloat
            {\hspace{-1.1mm}\vspace{0mm}\input{./fig/motivation/kitti.tex}}
  \end{minipage}%
  \begin{minipage}[t]{0.4467\columnwidth}
  \centering
  \subfloat
          {\hspace{-0.9mm}\vspace{0.18mm}\input{./fig/motivation/stara_voda.tex}}
  \end{minipage}
  \vspace{-5.5mm}
  \caption{\hlr{%
    A fast and noise-filtering 3D point cloud sampling can speed up real-time estimation pipelines.
    (a) An example of a single-frame sampling at the crossroad highlighted in (b) by each of the given methods (input point cloud in black).
    (b--c) Although sampling in the input space is uninformed about point-map correspondences, such sampling can improve performance if the sampling is fast and preserves the quality of the points (e.g., removes non-informative points).
    (b) Trajectory estimated by KISS-ICP~\cite{vizzo2023KISSICPDefensePointtoPoint} odometry on KITTI seq. \#00 when preceded by one of the sampling methods (ground truth in black).
    Similarly, (c) shows trajectory estimated on-board a \acs{mav} using LOAM~\cite{zhang2014LOAMLidarOdometry} odometry in a vertically self-symmetrical church in Star\'{a} Voda~\cite{petracek2023NewEraCultural}.}
    In LOAM, the plane and line features are sampled instead of the points.%
    }
  \label{fig:motivation}
  \vspace{-5mm}
\end{figure}

\hlr{
  The typical uninformed point cloud sampling methods include feature extraction~\cite{zhang2014LOAMLidarOdometry,qi2017PointNetDeepHierarchicala}, point-density normalization~\cite{rusu20113DHerePoint,vizzo2023KISSICPDefensePointtoPoint}, normal-space sampling~\cite{rusinkiewicz2001EfficientVariantsICP, gelfand2003GeometricallyStableSampling}, and learning-based inference~\cite{lang2020SampleNetDifferentiablePoint, nezhadarya2020AdaptiveHierarchicalDownSampling, wang2019DynamicGraphCNN, yang2018FoldingNetPointClouda, shen2018MiningPointCloud}.
  With the individual advantages and disadvantages of these widely used methodologies, the overall challenges remain in their effectiveness, latency minimization, and environment adaptability.
  The experimental part of this paper shows that finding optimal parameters of such methods is often a balance between speed and accuracy.
  Our analyses also show that the optimal parameters are rarely adaptable to different \acs{lidar} sensors, estimation pipelines, and environment types; and need to be exhaustively tuned for every instance.}

\hlr{
  Lastly, fast and noise-removing uniformed sampling has been shown to improve the performance of real-time pipelines in well-conditioned settings~\cite{vizzo2023KISSICPDefensePointtoPoint,zhang2014LOAMLidarOdometry}.
  However, it has also been shown that uninformed methods may improve performance in environments with a low amount of salient geometrical structures if these salient structures are part of the sampled data.
  We denote these settings, where the point cloud contains only a handful of exploitable structures, \textit{weakly} degenerate.
  These settings most notably emerge in geometrically symmetrical environments, such as subterranean tunnels~\cite{ozaslan2017AutonomousNavigationMapping} and caves~\cite{tranzatto2022CERBERUSDARPASubterranean}, and vertically-symmetrical historical monuments~\cite{petracek2023NewEraCultural}.}
\section{Related work}
\label{sec:sota}
\vspace{-0.5mm}


A ubiquitous point cloud sampling method is uniform sampling (voxelization), which discretizes space into fixed-sized cubes, each containing $N$ (typically 1) points at maximum.
Typical voxel-filter implementations employ an octree structure~\cite{schnabel2006OctreebasedPointcloudCompression} or use a simple numerical discretization, such as that implemented in \acs{pcl}~\cite{rusu20113DHerePoint}.
The feature extraction methods consist of learning-based solutions (such as PointNet++~\cite{qi2017PointNetDeepHierarchicala}) and hand-crafted feature (most commonly plane and line features defined in \acs{loam}~\cite{zhang2014LOAMLidarOdometry}) extractors.
\hlr{Although these methods perform reasonably well in geometrically rich settings when tuned properly, they are sensitive to parametrization.
Moreover, learning-based methods lack sampling guarantees and require each environment to be part of the training data.}

\hlr{%
A deterministic sampling method~\cite{rusinkiewicz2001EfficientVariantsICP} selects points such that their normals uniformly fill the normal-vector space.
The covariance-based sampling (CovS)~\cite{gelfand2003GeometricallyStableSampling} iteratively selects the points, which maximize the expected normal-based contribution to the \acs{dof} least constrained in the eigenspace of the sampled set.
Both methods \cite{gelfand2003GeometricallyStableSampling,rusinkiewicz2001EfficientVariantsICP} have shown that point normals can be a helpful mechanism in guiding the sampling under \textit{weak} geometrical degeneracy. 
However, obtaining the point normals cheaply, correctly, and reliably is challenging, especially given the projection nature of modern 3D \acp{lidar} that generate data with uneven density and surface occlusions.
}
PFilter~\cite{duan2022PFilterBuildingPersistenta} and ROI-cloud~\cite{zhou2020ROIcloudKeyRegion} are designed for use in a robot ego-motion estimation by employing previous \acs{lidar} scans.
PFilter~\cite{duan2022PFilterBuildingPersistenta} assigns each point a \textit{persistency-index}, quantifying how persistent the point is over a short history of measurements.
Static points, favorable in correspondence matching, tend to score higher in persistence.
The ROI-cloud~\cite{zhou2020ROIcloudKeyRegion} divides space into cubes weighted by the amount of inlying geometrical features.
\cite{zhou2020ROIcloudKeyRegion} then propagates virtual particles representing past measurements and fuses them with the weighted cubes.
Points are then sampled in areas where the weighted cubes align with the particles' distribution.

Among data-based sampling methods lies SampleNet~\cite{lang2020SampleNetDifferentiablePoint}, which learns task-specific sampling for object classification and geometry reconstruction.
The method in \cite{nezhadarya2020AdaptiveHierarchicalDownSampling} learns features and selects the points with the greatest contribution to the global max-pooling.
DGCNN~\cite{wang2019DynamicGraphCNN}, FoldingNET~\cite{yang2018FoldingNetPointClouda}, and KCNET~\cite{shen2018MiningPointCloud}
convert the point cloud into a graph and resample based on graph-based max-pooling, which takes the maximum features over the neighborhood of each vertex using a pre-built k-NN graph.
The disadvantage of these methods is the absence of deterministic guarantees that the sampling will be invariant to the type of environment, and that it will maximize point relevancy in estimation. 

Among the relevant redundancy-minimizing methods is~\cite{lim2022SingleCorrespondenceEnougha}.
Therein, the authors show that fewer correspondences are better in global registration, given that the correspondences are accurate.
A map-compressing method~\cite{li2023ReducingRedundancyMaps} then applies concepts of feature similarity to select only one of the nearby features, marking the rest redundant and removing them.
However, being formulated for expensive global registration and map compression, \cite{lim2022SingleCorrespondenceEnougha,li2023ReducingRedundancyMaps} are inapplicable in front end of a real-time ego-motion estimation of a robot.

It has also been proposed that sampling is to be performed at the optimization level once the point-to-map correspondences are found.
The greedy-based method~\cite{jiao2021GreedyBasedFeatureSelection} selects the optimization residuals such that the log-determinant of the approximate Hessian of the optimization problem is maximized.
KFS-LIO~\cite{li2021KFSLIOKeyFeatureSelection} does so similarly, but maximizes the inverse trace of the Hessian.
\mbox{X-ICP}~\cite{tuna2024XICPLocalizabilityAwareLiDAR} filters out residuals with non-parallel plane normals per each \acs{dof}, similarly to the normal-space equalization proposed in~\cite{rusinkiewicz2001EfficientVariantsICP}.
\hlr{%
Simplified version Xs-ICP~\cite{tuna2024XICPLocalizabilityAwareLiDAR} does similarly to X-ICP but reuses the residuals computed in the first iteration as a prior in subsequent iterations.
The advantage of sampling at the optimization level (informed) is the possibility to relate to the information theory, allowing to formulate awareness to degeneracy in the optimization.
In particular, \cite{jiao2021GreedyBasedFeatureSelection, li2021KFSLIOKeyFeatureSelection, tuna2024XICPLocalizabilityAwareLiDAR} utilize the eigenspace of the information matrix to quantify the degeneracy in the optimization, as introduced in~\cite{zhang2016DegeneracyOptimizationbasedState}.
However, residual-space sampling is sensitive to noise in correspondences and variability in point density and comes at a cost of re-sampling in every iteration of an estimation pipeline (see~\autoref{fig:estimation_pipeline}).
Uninformed input-space sampling is computed only once per point cloud, but cannot directly relate to the degeneracy without the correspondence pairings.}

\textbf{The contributions of this paper} include \hlr{a novel out-of-the-loop 3D point cloud sampling named Redundancy-Minimizing Sampling (RMS).
The method minimizes point redundancy} within a point cloud by maximizing the entropy of the \textit{gradient flow} in the sampled set.
\hlr{It} builds upon the fact that hyperplane surfaces (i.e., linear and planar surfaces) contain a high level of redundancy propagated into the iterative estimation pipeline.
Instead of classifying points into surface types, we propose a \ac{gfh} quantifying the potential of a point to lie on a hyperplane surface based on its local point distribution.
The method \hlr{is fast, uninformed, and deterministic} and does not need point-normals to be known, is independent on the environment, is effectively parametrizable by a single parameter only, and is integrable into most state-of-the-art \acs{lidar}-based odometries and \acsp{slam}, both \textit{dense} (using entire point clouds) and \textit{feature}-based.

\vspace{-2mm}
\section{Problem Definition}
\label{sec:problem_definition}
\vspace{-0.5mm}

The underlying problem of a six \acs{dof} robot ego-motion estimation from \acs{lidar} data is scan matching.
Scan matching can be formulated as finding the parameters $\theta^* \in SE(3)$, minimizing the squared sum of the residual functions $\vect{r} \in \mathbb{R}^3$ over two point sets $\mathcal{P} = \{\vect{p} \in \mathbb{R}^3\}$ and $\mathcal{Q} = \{\vect{q} \in \mathbb{R}^3\}$
\begin{equation}
  \argmin_{\theta \in SE(3)} \, g_{\theta}(\mathcal{P}, \mathcal{Q}) =
  \argmin_{\theta \in SE(3)} \, \sumC{\mathcal{P}}{\mathcal{Q}} \rho\left(||\vect{r}(\theta, \vect{p}, \vect{q})||_2^2\right),
  \label{eq:vanilla_opt}
  \vspace{-1mm}
\end{equation}
where $\mathbb{C}_{\mathcal{Q}}^{\mathcal{P}}$ represents the set of correspondence pairs from $\mathcal{P}$ to $\mathcal{Q}$ and $\rho$ is a robust kernel with outlier rejection properties.
Formulated as a pose estimation task, $\theta = \{\mathbf{t}, \mathbf{R}\}$ consists of a translation $\mathbf{t} \in \mathbb{R}^3$ and a rotation $\mathbf{R} \in SO(3)$ of the pose change from $\mathcal{P}$ to $\mathcal{Q}$.
Note that, $\mathcal{P}$ and $\mathcal{Q}$ can be entire \acs{lidar} scans in \textit{dense} or extracted features in \textit{feature}-based formulations, and that the most prevalent $\mathbf{r}$ functions in common iterative scan matchers are the \textit{point-to-point}, \textit{point-to-plane}, and \textit{point-to-line} metrics, \hlg{which are for a pair $\left(\mathbf{p}, \mathbf{q}\right) \in \mathbb{C}_{\mathcal{Q}}^\mathcal{P}$ given as}
\par{\eqsize\vspace{-3mm}
\begin{align}
  \hspace{0mm}
  \mathbf{r}^{\bullet} = \theta\mathbf{p} - \mathbf{q},\hspace{4mm}
  \mathbf{r}^{\square} = \left( \mathbf{n}\tran\mathbf{r}^{\bullet} + d \right)\mathbf{n},\hspace{4mm}
  \mathbf{r}^{|} = \mathbf{r}^{\bullet} - \left( \left(\mathbf{r}^{\bullet}\right)\tran \mathbf{v} \right)\mathbf{v},
  \label{eq:metrics}
  \vspace{-2mm}
\end{align}}%
where $(\mathbf{n},\,d)$ is the parametrization of a plane that $\mathbf{q}$ lies on ($\mathbf{n}$ is a unit surface normal), $\mathbf{v}$ is a unit direction of a line that $\mathbf{q}$ lies on, and $\theta\mathbf{p} = \mathbf{R}\mathbf{p} + \mathbf{t}$.

In the related correspondence selection methods~\cite{tuna2024XICPLocalizabilityAwareLiDAR, li2021KFSLIOKeyFeatureSelection,jiao2021GreedyBasedFeatureSelection}, the selection is formulated as finding a minimum-information correspondence subset that improves the performance of an iterative matching process in degenerate scenarios.
Commonly, these works formulate the problem as a minimization task
\par{\eqsize  
\vspace{-4mm}
\begin{align}
  &\minimize_{\theta \in SE(3)} \, \sum_{(\mathbf{p}, \mathbf{q}) \in \bar{\mathbb{C}}_{\mathcal{Q}}^{\mathcal{P}}}\rho\left(||\mathbf{r}(\theta, \mathbf{p}, \mathbf{q})||_2^2\right),\label{eq:sota_opt}\\
  &\text{subject to} \quad \bar{\mathbb{C}}_{\mathcal{Q}}^{\mathcal{P}} \subseteq \mathbb{C}_{\mathcal{Q}}^{\mathcal{P}},\,\bar{\mathbb{C}}_{\mathcal{Q}}^{\mathcal{P}} \neq \emptyset,
  \vspace{-2mm}
\end{align}}%
where $\bar{\mathbb{C}}_{\mathcal{Q}}^{\mathcal{P}}$ is a fixed-cardinality subset of correspondences selected from $\mathbb{C}_{\mathcal{Q}}^{\mathcal{P}}$ with respect to the log determinant~\cite{jiao2021GreedyBasedFeatureSelection} or inverse trace~\cite{li2021KFSLIOKeyFeatureSelection} of the information matrix, and as a sum of constraints per optimization direction in the objective function~\cite{tuna2024XICPLocalizabilityAwareLiDAR}.
Finding point-map correspondences \hlr{$\mathbb{C}_{\mathcal{Q}}^{\mathcal{P}}$} and then identifying the \hlr{optimal subset $\bar{\mathbb{C}}_{\mathcal{Q}}^{\mathcal{P}}$} is expensive, especially when repeatedly computed within iterative algorithms.

Proposed formulation decreases the problem dimensionality by selecting points in the input scan $\mathcal{P}$ before the iterative process of correspondence search, linearization, residual sampling, and optimization.
We formulate the pose estimation as
\par{\eqsize  
\vspace{-3mm}
\begin{align}
  \argmin_{\substack{\theta \in SE(3),\\\bar{\mathcal{P}} \subseteq \mathcal{P}}} g_{\theta}(\bar{\mathcal{P}}, \mathcal{Q}) = 
  \argmin_{\substack{\theta \in SE(3),\\\bar{\mathcal{P}} \subseteq \mathcal{P}}} \, \sumC{\bar{\mathcal{P}}}{\mathcal{Q}} \rho\left(||\mathbf{r}_{\theta}(\mathbf{p}, \mathbf{q})||_2^2 \right),\label{eq:proposed_opt_start}
\vspace{-2mm}
\end{align}}%
where $\mathbb{C}_{\mathcal{Q}}^{\bar{\mathcal{P}}}$ is a set of correspondence pairs from $\bar{\mathcal{P}}$ to $\mathcal{Q}$ and
\par
\begin{figure}[t]
  \vspace{-0mm}
  \centering
  \input{./fig/architecture.tex}
  \vspace{-2mm}
  \caption{
    Pipeline of an iterative pose estimation pipeline extended with (a) in-the-loop residual sampling and (b) a single-shot input data sampling.
    (a) The formulation (used in~\cite{jiao2021GreedyBasedFeatureSelection, li2021KFSLIOKeyFeatureSelection, tuna2024XICPLocalizabilityAwareLiDAR}) utilizes the full point cloud~$\set{P}$ and introduces a significant overhead in each iteration.
    (b) The proposed architecture includes a single-shot out-of-the-loop sampling, which lowers the overall complexity by reducing the input size.}
  \label{fig:estimation_pipeline}
  \vspace{-5mm}
\end{figure}
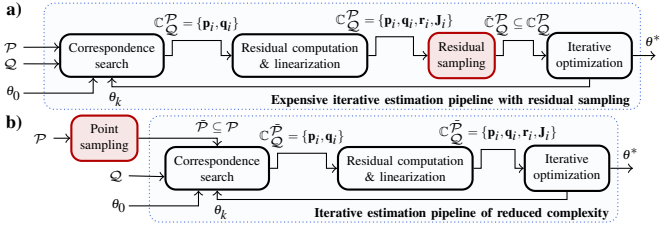

{\eqsize  
\vspace{-3mm}
\begin{align}
  &\bar{\mathcal{P}} = \argmin_{\Omega \in \{ \Theta\,|\,\Theta \subseteq \mathcal{P}, \Theta \neq \emptyset \}} |\Omega|,\\
  \text{subject to }\quad &\argmin_{\theta \in SE(3)} \, g_{\theta}(\mathcal{P}, \mathcal{Q}) = \argmin_{\theta \in SE(3), \bar{\mathcal{P}} \subseteq \mathcal{P}} \, g_{\theta}(\bar{\mathcal{P}}, \mathcal{Q}).
  \label{eq:proposed_opt_end}
\vspace{-1mm}
\end{align}}%
In other words, we formulate the problem as finding a minimum-cardinality subset $\bar{\mathcal{P}} \subseteq \mathcal{P}$ over which the minimization problem converges to the same optimum as in the original formulation.
Differences in iterative pipelines using formulations in Eq.~\eqref{eq:sota_opt} and Eq.~\eqref{eq:proposed_opt_start} are shown in \autoref{fig:estimation_pipeline}.


\vspace{-1mm}
\section{Information Redundancy Minimization}
\label{sec:inf_redund_min}

The problem formulated in Eq. \eqref{eq:proposed_opt_start}--\eqref{eq:proposed_opt_end} requires finding a minimal-cardinality non-empty subset of points $\bar{\mathcal{P}} \subseteq \mathcal{P}$ over which the estimation converges to the optimum without prior knowledge about the correspondences among point sets $\mathcal{P}$~and~$\mathcal{Q}$.
This makes the formulation NP-hard and non-causal as the information about a point contribution to the optimization is unknown without its target correspondence.
When a correspondence 
$\left( \mathbf{p}_i, \mathbf{q}_i \right)$
is known, the related works~\cite{tuna2024XICPLocalizabilityAwareLiDAR, li2021KFSLIOKeyFeatureSelection,jiao2021GreedyBasedFeatureSelection} define its contribution in relation to the eigenspace of the information matrix $\leftscript{i}{\mathbf{J}}_{\theta}\tran\leftscript{i}{\mathbf{J}}_{\theta}$,
where 
\begin{equation}
  \leftscript{i}{\mathbf{J}}_{\theta} = \left[ \frac{\partial \mathbf{r}_{\theta}\left(\mathbf{p}_i,\,\mathbf{q}_i \right)}{\partial \mathbf{t}},\,\frac{\partial \mathbf{r}_{\theta}\left(\mathbf{p}_i,\,\mathbf{q}_i \right)}{\partial \mathbf{R}}\right]
  \label{eq:jacobian_definition}
\end{equation}
is the Jacobian of the residual function $\mathbf{r}_{\theta}$ (e.g., $\mathbf{r}_{\theta}^{\square}$ from Eq.~\eqref{eq:metrics} in~\cite{jiao2021GreedyBasedFeatureSelection}), or with relation to the approximate Hessian of the opt. problem~\cite{zhang2016DegeneracyOptimizationbasedState} given as $(\leftscript{\mathcal{P}}{\mathbf{J}}_{\theta})\tran\leftscript{\mathcal{P}}{\mathbf{J}}_{\theta}$, where $\leftscript{\mathcal{P}}{\mathbf{J}}_{\theta} = \sum_{i=1}^{|\mathbb{C}_{\mathcal{Q}}^{\mathcal{P}}|}\,\leftscript{i}{\mathbf{J}}_{\theta}$.
Although this makes the problem causal, finding the optimal minimum-cardinality subset is still NP-hard; and remains an open challenge.

To tackle this problem, we propose to approximate the solution to the problem formulated in Eq. \eqref{eq:proposed_opt_start}--\eqref{eq:proposed_opt_end} by defining, finding, and removing redundancy within a point set without knowledge about the correspondences.
When applied to a typical iterative process of a robot's ego-motion estimation, the proposed solution inherently removes noise and lowers the computational latency.
When under real-time termination criteria (e.g., number of iterations, rate of change), the lowered cost improves the rate and accuracy of convergence.
\vspace{-2mm}
\subsection{Redundancy in a Point Set}
\label{sec:redundancy_in_a_point_set}


Every perceived environment can be decomposed into a set of $S$ atomic surfaces $\mathbb{S} = \bigcup_{s \in \left<1, S\right>}\mathbb{S}_s$ of arbitrary complexities, ranging from linear and planar to quadratic and other nonlinear areas.
In this work, the environment is assumed to be decomposable into linear and planar (hyperplane) surfaces.
An input point set $\mathcal{P}$ can then be understood as a discretization of the observed hyperplane surfaces $\mathcal{P} = \bigcup_{s \in \left<1, S\right>}\mathcal{P}_s$, where $\mathcal{P}_s$ represents a set of points observed on the surface $\mathbb{S}_s$.

\begin{definition}
A single-surface point set $\mathcal{P}_s$ contains information redundancy if removing one or multiple points from the set does not change its rate of information (average entropy) regarding the optimization problem.
\label{def:redundancy_definition}
\end{definition}
\begin{remark}
In the optimization task defined in Eq.~\eqref{eq:vanilla_opt}-\eqref{eq:proposed_opt_start}, the redundancy represents points that generate identical (parallel and of the same magnitude) residuals, whose removal does not alter the loss function, nor does it change the global optimum of the objective function.
\end{remark}

Without the loss of generality, the robust kernel in Eq.~\eqref{eq:vanilla_opt}-\eqref{eq:proposed_opt_start} can be omitted for now, and a set-residual function (the objective function) can be defined as the sum of the point residuals
\begin{equation}
  \mathbf{r}_{\theta}(\mathcal{P}) = \sumC{\mathcal{P}}{\mathcal{Q}} ||\mathbf{r}_{\theta}(\mathbf{p}, \mathbf{q})||_2^2
  \label{eq:set_residual}
\vspace{-1mm}
\end{equation}
to be minimized.
Given the set of atomic surfaces $\mathbb{S}$ and their corresponding point sets $\mathcal{P}_s$, Eq.~\eqref{eq:set_residual} can be equivalently defined as a sum of surface-subset residuals
\begin{equation}
  \mathbf{r}_{\theta}(\mathbb{S}) = 
  \sum_{\mathcal{P}_s \in \mathbb{S}} 
  \sumC{\mathcal{P}_s}{\mathcal{Q}} ||\mathbf{r}_{\theta}(\mathbf{p}, \mathbf{q})||_2^2.
  \label{eq:surface_subset_residual}
\vspace{-1mm}
\end{equation}
\begin{definition}
  Without altering the \hlr{translational optimum}, the objective function can be defined as a sum of set-residual rates
  \begin{equation}
    \bar{\mathbf{r}}_{\theta}(\mathbb{S}) 
    = \sum_{\mathcal{P}_s \in \mathbb{S}}\overline{\mathbf{r}}_{\theta}(\mathcal{P}_s)
    = \sum_{\mathcal{P}_s \in \mathbb{S}}\dfrac{1}{\left|\hlr{\mathbb{C}^{\mathcal{P}_s}_{\mathcal{Q}}}\right|}
      \sumC{\mathcal{P}_s}{\mathcal{Q}}
      ||\mathbf{r}_{\theta}(\mathbf{p}, \mathbf{q})||_2^2.
    \label{eq:setresidual_rates}
    \vspace{-2mm}
  \end{equation}
  \label{def:obj_func}
\end{definition}

\vspace{-4mm}
\begin{proof}
  The Jacobian of the \hlr{obj.} function defined in Eq.~\eqref{eq:set_residual} is
\par{\eqsize  
\vspace{-3mm}
\begin{align}
  \leftscript{\mathcal{P}}{\mathbf{J}}_{\theta}
  &= \frac{\partial \left( \sum_{(\mathbf{p}, \mathbf{q}) \in \mathbb{C}_{\mathcal{Q}}^{\mathcal{P}}}
       ||\mathbf{r}_{\theta}(\mathbf{p}, \mathbf{q})||_2^2  \right)}{\partial \theta}\nonumber\\
  &= \sumC{\mathcal{P}}{\mathcal{Q}} \frac{\partial ||\mathbf{r}_{\theta}(\mathbf{p}, \mathbf{q})||_2^2 }{\partial \theta}
   = 2\sumC{\mathcal{P}}{\mathcal{Q}} \mathbf{r}_{\theta}(\mathbf{p}, \mathbf{q})
   \label{eq:jacobian_vanilla}
\end{align}
}%
and the Jacobian of Eq.~\eqref{eq:surface_subset_residual} is given as
\par{\eqsize  
\vspace{-3mm}
\begin{align}
  \leftscript{\mathbb{S}}{\mathbf{J}}_{\theta}
  &= \frac{\partial \left( \sum_{\mathcal{P}_s \in \mathbb{S}} \sumC{\mathcal{P}_s}{\mathcal{Q}} ||\mathbf{r}_{\theta}(\mathbf{p}, \mathbf{q})||_2^2  \right)}{\partial \theta}\nonumber\\
  &= \sum_{\mathcal{P}_s \in \mathbb{S}} |\mathcal{P}_s| \frac{\partial  ||\leftscript{s}{\mathbf{r}}_{\theta}(\mathbf{p}, \mathbf{q})||_2^2 }{\partial \theta}
  = 2\sum_{\mathcal{P}_s \in \mathbb{S}} |\mathcal{P}_s| \, \leftscript{s}{\mathbf{r}}_{\theta}(\mathbf{p}, \mathbf{q}),
  \label{eq:jacobian_surface_subset_residual}
\end{align}
\vspace{-1mm}}%
where $\leftscript{s}{\mathbf{r}}_{\theta}$ is a common residual for the redundant surface $s$.
As each hyperplane surface $s$ contains $|\mathcal{P}_s|$ identical residuals \hlr{(see Def.~\ref{def:pminus1_redundant_residuals})}, the simplification
$\sumC{\mathcal{P}_s}{\mathcal{Q}}||\mathbf{r}_{\theta}(\mathbf{p}, \mathbf{q})||_2^2 = |\mathcal{P}_s| \cdot ||\leftscript{s}{\mathbf{r}}_{\theta}||_2^2$
makes the Jacobians in Eq.~\eqref{eq:jacobian_vanilla} and Eq.~\eqref{eq:jacobian_surface_subset_residual} identical, assuming perfect point-to-surface associations.
The Jacobian of Eq.~\eqref{eq:setresidual_rates} is derived similarly \hlr{as in Eq.~\eqref{eq:jacobian_vanilla} and \eqref{eq:jacobian_surface_subset_residual}} as
\par{\eqsize  
\vspace{-3mm}
\begin{align}
  \leftscript{\mathbb{S}}{\bar{\mathbf{J}}}_{\theta} &= 
  \frac{\partial \left( \sum_{\mathcal{P}_s \in \mathbb{S}} \bar{\mathbf{r}}_{\theta}(\mathcal{P}_s) \right)}{\partial \theta}
  = \sum_{\mathcal{P}_s \in \mathbb{S}} \frac{1}{|\mathcal{P}_s|} \frac{\partial \left( \sumC{\mathcal{P}_s}{\mathcal{Q}} ||\mathbf{r}_{\theta}(\mathbf{p}, \mathbf{q})||_2^2 \right)}{\partial \theta}\nonumber\\
  &= \sum_{\mathcal{P}_s \in \mathbb{S}} \frac{1}{|\mathcal{P}_s|} \frac{\partial \left( |\mathcal{P}_s|\cdot||\leftscript{s}{\mathbf{r}}_{\theta}||_2^2 \right)}{\partial \theta}
   = 2\sum_{\mathcal{P}_s \in \mathbb{S}} \leftscript{s}{\mathbf{r}}_{\theta}.
   \label{eq:problem_jacobian_rate}
\end{align}
}%
The Hessian matrices of all three formulations are \hlr{given as}
\par{\eqsize  
\vspace{-3mm}
\begin{align}
  \leftscript{\mathcal{P}}{\mathbf{H}}_{\theta} = 2\sum_{i = 1}^{|\mathcal{P}|} \leftscript{i}{\mathbf{J}}_{\theta},\hspace{5mm}
  \leftscript{\mathbb{S}}{\mathbf{H}}_{\theta} = 2\sum_{s = 1}^{|\mathcal{P}_s|} |\mathcal{P}_s| \, \leftscript{s}{\mathbf{J}}_{\theta},\hspace{5mm}
  \leftscript{\mathbb{S}}{\bar{\mathbf{H}}}_{\theta} = 2\sum_{s = 1}^{|\mathcal{P}_s|} \leftscript{s}{\mathbf{J}}_{\theta},
  \label{eq:hessian_3}
\end{align}
}%
where $\leftscript{s}{\mathbf{J}}_{\theta}$ is the Jacobian of $\leftscript{s}{\mathbf{r}}_{\theta}$, as per \hlr{Eq.~\eqref{eq:analytical_jacobian_point}--\eqref{eq:analytical_jacobian_line} below}.

The Jacobians 
of the residual functions \hlr{(defined in Eq.~\eqref{eq:metrics})} are given analytically \hlr{according to Eq.~\eqref{eq:jacobian_definition}} as
\par{\eqsize  
\vspace{-3mm}
\begin{align}
  \leftscript{i}{\mathbf{J}}_{\theta}^{\bullet} &=
  \left[
    \mathbf{I},\,\,-\mathbf{R}[\mathbf{p}_i]_{\times}
    \right],
  \label{eq:analytical_jacobian_point}
  \\
  \leftscript{i}{\mathbf{J}}_{\theta}^{\square} &= 
  \left[
    \mathbf{n}\tran_i\mathbf{n}_i,\,\,-\mathbf{n}_i\tran\mathbf{n}_i\mathbf{R}[\mathbf{p}_i]_{\times}
    \right],
  \label{eq:analytical_jacobian_plane}
  \\
  \leftscript{i}{\mathbf{J}}_{\theta}^{|} &= 
  \left[
    \mathbf{I} - \mathbf{v}_i\mathbf{v}_i\tran,\,\,-\left(\mathbf{I} - \mathbf{v}_i\mathbf{v}_i\tran\right)\mathbf{R}[\mathbf{p}_i]_{\times}
    \right],
  \label{eq:analytical_jacobian_line}
\end{align}}%
given that $\mathbf{I} \in \mathbb{R}^{3 \times 3}$ and $\frac{\partial}{\partial\mathbf{R}}\left(\mathbf{R}\mathbf{p}_i\right) = -\mathbf{R}[\mathbf{p}_i]_{\times}$, where $[\mathbf{p}_i]_{\times} \in \mathbb{R}^{3\times 3}$ is the skew-symmetric matrix of $\mathbf{p}_i$.
\hlr{%
It is clear that in the translational space (the ${\partial}/{\partial}\mathbf{t}$ part of the Jacobians), the change in residuals is, for the most common metrics, either constant or a function of the surface parameters.
Since the translational change depends only on the surface $s$, selecting a single residual $\mathbf{r}_s$ per surface preserves the basis of the translational eigenspace of both the Jacobian and Hessian matrices.
Thus, the global optimum in the translational space of the objective function remains unchanged.}
\end{proof}

\vspace{-1mm}
\begin{remark}
  Although this reformulation does not alter the \hlr{translational} optimum, it reshapes the \hlr{respective part of the objective func.} $g_{\theta}$ to $\bar{g}_{\theta}$ without changing its monotonic intervals
  \begin{equation}
    \forall \mathbf{x}, \mathbf{y} \in \mathbb{R}^D,\quad g_{\theta}(\mathbf{x}) \odot g_{\theta}(\mathbf{y}) \Rightarrow \bar{g}_{\theta}(\mathbf{x}) \odot \bar{g}_{\theta}(\mathbf{y}),
    \label{eq:monotonic_intervals}
  \end{equation}
  where $\odot$ is any linear inequality operator and $D$ is the problem dimensionality.
\end{remark}

\vspace{1mm}
\begin{remark}
\hlr{%
When constrained to an ego-motion estimation task, we can assume the rotation changes to be small.
Under this assumption, the first-order linearization of $\mathbf{R}$ is given as $\mathbf{R} \approx \mathbf{I} + [\mathbf{b}]_{\times}$, where $\mathbf{b} = [\alpha, \beta, \gamma]\tran$ is vector of the three rotational \acp{dof}.
Then, the rotational space in Eq.~\eqref{eq:analytical_jacobian_point}--\eqref{eq:analytical_jacobian_line} reduces to a function of surface parameters and $[\mathbf{p}_i]_{\times}$, which denotes that sensitivity to rotations increases with point distance.
This means that two points are also redundant in the rotational space if they belong to the same surface and have equal $[\mathbf{p}_i]_{\times}$.}
\label{remark:small_angles_approximation}
\end{remark}


\begin{definition}
  Assuming zero noise, every hyperplane surface $\mathbb{S}_s$ generates $|\mathcal{P}_s| - 1$ redundant residuals.
\label{def:pminus1_redundant_residuals}
\end{definition}
\begin{proof}
  Given the hyperplane surfaces and their point-set observations $\mathcal{P}_s$, the set-residual rate
  $\bar{\mathbf{r}}_{\theta}(\mathcal{P}_s) = \bar{\mathbf{r}}_{\theta}(\Pi)$
  applies for all $\Pi \in \{ \pi \, | \, \pi \subset \mathcal{P}_s, \pi \neq \emptyset \}$.
  Eq.~\eqref{eq:problem_jacobian_rate} \hlr{then} shows that reducing the cardinality of $\set{P}_s$ from $|\set{P}_s|$ to $1$ preserves the \hlr{translational} optimum, which implies that $|\set{P}_s| - 1$ residuals are redundant.
\end{proof}

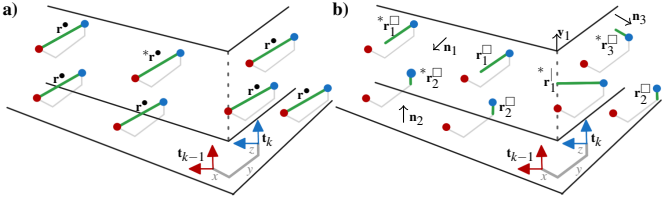
\begin{figure}
  \centering
  \vspace{0mm}
  \input{./fig/redundancy.tex}
  \vspace{-5mm}
  \caption{
    Simplistic case of point redundancy (Def.~\ref{def:obj_func} and~\ref{def:pminus1_redundant_residuals}) for a robot translating from the position $\mathbf{t}_{k-1}$ (\tikzcircle[color_red, fill=color_red]{2.0pt} points) to $\mathbf{t}_k$ (\tikzcircle[RoyalBlue, fill=RoyalBlue]{2.0pt} points).
    (a) The \textit{point-to-point} metric generates identical residuals, which makes the residual rate constant for any positive number of residuals used, e.g., a single residual generates
        $\bar{\mathbf{r}}(\mathcal{P}) = \leftscript{*}{\mathbf{r}^{\bullet}}$.
    (b) The point-to-hyperplane metrics generate identical residuals per surface (in this example, the surfaces comprise three planes and a single line).
        The minimized objective function remains constant if any positive number of points is sampled per each surface.
        In this case, using the minimum amount of samples yields the residual rate $\bar{\mathbf{r}}(\mathcal{P}) = \leftscript{*}{\mathbf{r}_1^{|}} + \sum_{i=1}^{3}\leftscript{*}{\mathbf{r}_i^{\square}}$.
        This example assumes perfect correspondences, which is unrealistic under noise and rotation.
        The point sampling method proposed in \autoref{sec:quantifying_redundancy} is designed to be robust to cases where this assumption is not met.
    }
  \label{fig:surfaces}
  \vspace{-4mm}
\end{figure}

\autoref{fig:surfaces} shows an idealized case demonstrating redundancy in surface-point sets, as defined in Def.~\ref{def:obj_func} and \ref{def:pminus1_redundant_residuals}.

\begin{definition}
  Assuming the presence of noise, Def.~\ref{def:obj_func} and \ref{def:pminus1_redundant_residuals} can be generalized to find the min-cardinality non-empty subset $\hat{\mathcal{P}}_s \subseteq \mathcal{P}_s$ whose set-residual rate matches the one of its superset
\par{\eqsize\vspace{-5mm}
\begin{align}
  &\hat{\mathcal{P}}_s = \argmin_{\Omega \in \{ \Theta \, | \, \Theta \subseteq \mathcal{P}_s, \Theta \neq \emptyset \}} |\Omega|
  \label{eq:minimizing_residual_redundancy_A}\\
  &\phantom{}\text{subject to} \quad \bar{\mathbf{r}}_{\theta}\left( \Omega \right) = \bar{\mathbf{r}}_{\theta} \left( \mathcal{P}_s \right),
  \label{eq:minimizing_residual_redundancy_B}
\end{align}}%
  for each set of surface points $\mathcal{P}_s$.
  Given this formulation, each surface contains $|\mathcal{P}_s \setminus \hat{\mathcal{P}}_s|$ redundant residuals.
  Substituting Eq.~\eqref{eq:minimizing_residual_redundancy_A} into Eq.~\eqref{eq:setresidual_rates} yields the obj. function in the form of
\par{\eqsize  
\vspace{-3mm}
\begin{align}
  \mathbf{r}_{\theta}(\mathbb{S}) = \sum_{\mathcal{P}_s \in \mathbb{S}}\overline{\mathbf{r}}_{\theta}(\hat{\mathcal{P}}_s).
  \label{eq:residual_with_subset}
\end{align}
}%
\label{def:redund_minimization_generalization}
\end{definition}
\vspace{-5mm}
\begin{remark}
The reformulation is feasible since the Def.~\ref{def:obj_func} maintains the convergence properties exploitable by the nonlinear solvers.
The optimum consistency further satisfies Eq.~\eqref{eq:proposed_opt_end}.
\end{remark}

In practice, the data are \hlr{usually unstructured and are} subjected to noise, making it expensive to segment the input point set $\mathcal{P}$ into a set of surfaces, even trivially.
Instead of finding and segmenting the underlying surfaces \hlr{(as defined in Def.~\ref{def:pminus1_redundant_residuals} and \ref{def:redund_minimization_generalization})}, we propose in \autoref{sec:quantifying_redundancy} a heuristic for the direct quantification of the redundancy without point-surface associations.
In \autoref{sec:removing_the_redundancy}, we then propose a redundancy-minimizing algorithm robust towards noise, independent of correspondence matching, and invariant to small rotations.

\vspace{-2mm}
\subsection{Quantifying the Redundancy}
\label{sec:quantifying_redundancy}

As discussed at the beginning of \autoref{sec:inf_redund_min}, our objective is to find redundancy within a point set $\mathcal{P}$ without knowing the correspondences $\mathbb{C}_{\mathcal{Q}}^{\mathcal{P}}$ beforehand.
We tackle this by introducing a \acl{gfh} quantifying the uniqueness of a point by local flow of a geometric gradient.
The \ac{gfh} maximizes the potential of points in bringing unique information to the optimization once their correspondences are found.
Instead of expensive segmentation of the set $\mathcal{P}$ into surface observations~$\mathcal{P}_s$ \hlr{(as formulated in Def.~\ref{def:pminus1_redundant_residuals} and \ref{def:redund_minimization_generalization})}, the \ac{gfh} quantifies whether a point is locally a part of any hyperplane.
Since generating multiple residuals on a single hyperplane is a source of the redundancy (as defined in Def.~\ref{def:redundancy_definition}--\ref{def:redund_minimization_generalization}), this opens a way to the redundancy minimization discussed in \autoref{sec:removing_the_redundancy}.

The \ac{gfh} emerges from Def.~\ref{def:obj_func} and \ref{def:pminus1_redundant_residuals}, which define that identical (in orientation and magnitude) residuals are redundant in structuring the objective function and that on a single hyperplane, the residuals are identical inherently.
To quantify the uniqueness of points (and thus, the potential of future residuals), the \ac{gfh} finds the neighbors of each point $\mathbf{p} \in \mathcal{P}$ within a spherical neighborhood with radius $\lambda_{\mathbf{p}}\,\left(\si{\meter}\right)$
\begin{equation}
  \hlr{%
  \set{N}_{\mathbf{p}} = \{ \mathbf{j} \,\, | \,\, || \mathbf{j} - \mathbf{p} ||_2 < \lambda_{\mathbf{p}}, \mathbf{j} \neq \mathbf{p}, \mathbf{j} \in \mathcal{P} \},}
  \label{eq:gh_neighborhood}
\end{equation}
\vspace{0mm}%
and defines the \textit{gradient flow} (in meters) as
\begin{equation}
  \hlr{%
  \Delta_{\mathbf{p}} = \frac{1}{|\set{N}_{\mathbf{p}}|}\sum_{\mathbf{j} \in \set{N}_{\mathbf{p}}}\mathbf{j} - \mathbf{p}.}
  \label{eq:gfh}
\end{equation}

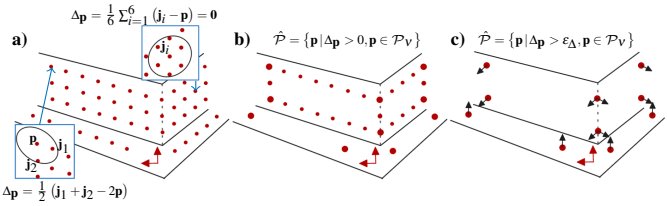
\begin{figure}[t]
  \centering
  \vspace{0mm}
  \hspace{-1mm}
  \input{./fig/gfh/gfh.tex}
  \vspace{-6mm}
  \caption{%
    Proposed \acl{gfh} for quantifying redundancy in a point set.
    (a) \ac{gfh} is computed for each point in a voxelized point set $\mathcal{P}_{\nu}$.
    (b) Points on the perceived borders have generally non-zero $\Delta_{\mathbf{p}}$, whereas (c) corner points yield the maximum $\Delta_{\mathbf{p}}$ (herein thresholded by an abstract value $\epsilon_{\Delta}$).
    Keeping the max-$\Delta_{\vect{p}}$ subset (c) ensures that all directions remain constrained, as shown by the black axes representing which translational directions the points constraint.}
  \label{fig:gh_and_redundancy_minimization}
  \vspace{-5mm}
\end{figure}

As demonstrated in \autoref{fig:gh_and_redundancy_minimization}, the \ac{gfh} scores high for points lying on the borders of a surface and low for points lying inside the borders (inliers).
Maximizing \ac{gfh} thus leads to prioritizing the borders of surfaces rather than the surface inliers, which is important for two reasons.
First, the borders in $\set{P}$ have the largest potential for correct correspondence matching with the borders of the corresponding physical surface.
Second, the inliers can generate erroneous local minima and resist sliding along the directions of a hyperplane when using the \textit{point-to-point} metric.
As discussed in~\cite{pottmann2003GeometrySquaredDistance}, the point-to-hyperplane metrics do not suffer from this deficiency, but it is still valuable to remove the redundancy to increase efficiency.

Every point-set matching algorithm more or less voxelizes the input set by a constant voxel size factor $\nu$ in order to reduce the cardinality of the input point set to $\mathcal{P}_{\nu} \subset \mathcal{P}$.
We exploit this by setting $\lambda_{\mathbf{p}} = 2\nu$ for unstructured point sets.
In structured point sets coming from a rotating 3D \ac{lidar} (e.g., Ouster), the neighborhood radius instead respects the projective properties of these sensors as
\begin{equation}
  \lambda_{\mathbf{p}} = 2 \max \left[ \nu, ||\mathbf{p}||_2 \max \left( \sin \frac{2\pi}{C},\, \sin \frac{\theta_v}{R - 1} \right) \right],
  \label{eq:neighborhood_radius}
\end{equation}
where $\theta_v$ is vertical and $2\pi$ is a horizontal field of view of the sensor, which data are generated in a matrix form with $R$ rows and $C$ columns.

The neighborhood search of Eq~\eqref{eq:gh_neighborhood} is the only expensive part of the proposed methodology.
We construct a KD-tree from the voxelized point set $\mathcal{P}_{\nu}$ to lower the cost.
Using $\mathcal{P}_{\nu}$ lowers the construction cost of the KD-tree and reduces the number of KD-tree queries to $|\mathcal{P}_{\nu}|$.
With construction complexity $\mathcal{O}\left(n\log{n}\right)$ and worst-case complexity of $n$-query radius search being $\mathcal{O}\left(n^2\log{n}\right)$ (where $n$ is $|\mathcal{P}|$ in the full and $|\mathcal{P}_{\nu}|$ in the voxelized case),
the overall complexity is reduced since $|\mathcal{P}_{\nu}| < |\mathcal{P}|$.
We show in \autoref{sec:results} that the overhead for computing the \ac{gfh} for all the points lowers the complexity of the ego-motion estimation and accelerates the full pipeline.


\vspace{-1mm}
\subsection{Removing the Redundancy}
\label{sec:removing_the_redundancy}

Although a redundancy might be beneficial for reducing the effects of noise and outliers, it makes the iterative process of correspondence finding, residual generation, linearization, and optimization more expensive.
Under the presence of termination criteria, the process may be undesirably hindered by accurate in-time convergence.


To find a solution to the NP-hard problem formulated in Eq.~\eqref{eq:minimizing_residual_redundancy_A}-\eqref{eq:residual_with_subset}, we could propose to solve an optimization task
\par{\eqsize  
\vspace{-3mm}
\begin{align}
  \hat{\mathcal{P}} = \argmin_{ \Omega \in \{ \Theta \,|\, \Theta \subseteq \mathcal{P}_{\nu}, \Theta \neq \emptyset \} } \Gamma_{\Delta}(\Omega),\quad
  \phantom{}\text{subject to} \quad |\Omega| = N_\Omega,
\end{align}}%
minimizing redundancy $\Gamma_{\Delta}$ in the \textit{gradient flow} of the subset $\Omega$ under a constraint on fixed cardinality of the output set $N_\Omega \in \left( 1, |\set{P}_\nu| \right>$. 
Although the concept of a fixed-cardinality constraint is common within the related works~\cite{li2021KFSLIOKeyFeatureSelection,jiao2021GreedyBasedFeatureSelection}, the notion of redundancy allows for a more rigorous formulation.
Since minimizing redundancy in data can be understood as maximizing the data entropy, we instead define a dual task
\par{\eqsize  
\vspace{-3mm}
\begin{align}
  \hat{\mathcal{P}} = \argmax_{ \Omega \in \{ \Theta \,|\, \Theta \subseteq \mathcal{P}_{\nu}, \Theta \neq \emptyset \} } H_{\Delta}(\Omega),\quad
  \phantom{}\text{subject to} \quad \bar{H}_{\Delta}(\Omega) \leq \lambda_{\bar{H}},
  \label{eq:maximizing_entropy}
\end{align}}%
maximizing the entropy of information $H_{\Delta}$ in the \textit{gradient flow} of the subset $\Omega$ under the termination criteria on the relative information rate $\bar{H}_{\Delta}$ (defined in Eq.~\eqref{eq:relative_entropy_rate}), given a maximum relative entropy rate $\lambda_{\bar{H}}\,\left( \si{\percent} \right)$.
The termination criteria in Eq.~\eqref{eq:maximizing_entropy} replaces the constraint on a fixed cardinality, which allows the selection to emergently adapt to the distribution of the points, making this method invariant to the type of environment.
By thresholding the relative information rate via $\lambda_{\bar{H}}$, a certain level of redundancy is introduced into the system, possibly increasing robustness towards noise and outliers.

Let an entropy rate be an average entropy per point in set $\Omega$
\begin{equation}
  \bar{\mu}_{\Delta}(\Omega) = \frac{1}{|\Omega|}H_{\Delta}(\Omega),
\end{equation}
where the entropy of the set is given as
\begin{equation}
  H_{\Delta}(\Omega) = -\sum_{\mathbf{p} \in \Omega} p(\Delta_{\mathbf{p}})\log{p(\Delta_{\mathbf{p}})},
\end{equation}
and $p$ represents the probability of observing the \ac{gfh} value $\Delta_{\mathbf{p}}$.
The relative entropy rate (conditioned in Eq.~\eqref{eq:maximizing_entropy}) is defined as the normalized entropy rate
\begin{equation}
  \bar{H}_{\Delta}(\Omega) = \frac{1}{\bar{\mu}^*_{\Delta}(\Omega)}\bar{\mu}_{\Delta}(\Omega),
  \label{eq:relative_entropy_rate}
  \vspace{-1mm}
\end{equation}
where
\begin{equation}
  \bar{\mu}^*_{\Delta}(\Omega) = \max_{ \Psi \in \{ \Theta \,|\, \Theta \subseteq \Omega, \Theta \neq \emptyset \} } \bar{\mu}_{\Delta}(\Psi) 
  \label{eq:maximum_entropy_rate}
\end{equation}
represents the maximum entropy rate of all non-empty subsets $\Psi \subseteq \Omega$.
Note that the entropy rate is an inverse function to redundancy within the data, which allows us to formulate the dual task in Eq.~\eqref{eq:maximizing_entropy}.

The probability function $p$ is a function of the data $\mathcal{P}_\nu$, which are a function of the environment.
To maintain invariance to the environment, $p$ can not be modeled with a probability density function.
Instead, we propose to use a frequentist's approach to approximate the probability function~$p$.
First, the \ac{gfh} of each point in $\mathcal{P}_{\nu}$ is converted to its normalized norm
\begin{equation}
  \bar{\Delta}_{\mathcal{P}_{\nu}} = \left\{ \frac{||\Delta_{\mathbf{p}}||_2}{\max ||\Delta_{\mathcal{P}_{\nu}}||_2} \,\middle\vert\, \mathbf{p} \in \mathcal{P}_{\nu} \right\},
  \label{eq:gfh_normalized_norm}
\end{equation}
where $\max||\mathcal{P}_{\nu}||_2$ represents the maximum $||\Delta_{\mathbf{p}}||_2$ of any point $\mathbf{p} \in \mathcal{P}_\nu$.
Second, a histogram $\mathcal{H}_\Delta$ with $K$ bins bounded in interval $\left< 0, 1\right>$ is created out of the normalized \ac{gfh} norms $\bar{\Delta}_{\mathcal{P}_\nu}$, where each bin $k \in \left<1, K\right>$ holds a point set $\leftscript{k}{\mathcal{H}_\Delta}$.
The probability of a bin $k$ is then approximated by $p_k = \frac{|\leftscript{k}{\mathcal{H}_\Delta}|}{|\mathcal{P}_\nu|}$.


In \autoref{alg:entropy_maximization}, we propose a point sampling routine following formulation in Eq.~\eqref{eq:maximizing_entropy}.
Given the fact that the uniform distribution function yields a maximum entropy, ~\autoref{alg:entropy_maximization} maximizes uniformity in the \acs{gfh} by normalizing \acs{gfh} values in histogram $\mathcal{H}_\Delta$.
The routine constructs an empty histogram $\hat{\mathcal{H}}_\Delta$ and iteratively moves points from $\mathcal{H}_\Delta$ to $\hat{\mathcal{H}}_\Delta$.
This is done by per-row sampling from $\mathcal{H}_\Delta$ via cyclic iterative selection, going from greater to lower bins and moving a single point in each of the bins $k$ (if there is any) to the corresponding bin $k$ in $\hat{\mathcal{H}}_\Delta$.
The primary and secondary keys of sampling from a bin $k$ are 
\par{\eqsize\vspace{-4mm}
\begin{align}
  \mathbf{p}_k = \argmax_{\mathbf{p} \in \leftscript{k}{\mathcal{H}_\Delta}} \Delta_\mathbf{p},\hspace{10mm}
  \mathbf{p}_k = \argmax_{\mathbf{p} \in \leftscript{k}{\mathcal{H}_\Delta}} ||\mathbf{p}||_2.
  \label{eq:sorting_keys}
\end{align}}%
\hlr{%
  RMS does not balance rotation-space observability but exploits the fact that the rotational rate of residuals is a function of $||\mathbf{p}||_2$, as defined in Rem.~\ref{remark:small_angles_approximation}.
  This is done via the secondary key in Eq.~\eqref{eq:sorting_keys}, which values points by their potential for being part of a large-magnitude residual in the later correspondence-matching part of the estimation.
  Note that when the assumption of small rotations in Rem.~\ref{remark:small_angles_approximation} is not met, the invariance to rotations no more applies, leading to suboptimal sampling.}

The iterative sampling process is terminated once the termination criteria in Eq.~\eqref{eq:maximizing_entropy} is satisfied and $i \geq K$.
\hlg{%
Since the entropy reaches its maximum at $K$ steps, the maximum entropy rate $\bar{\mu}^*_{\Delta}$ is guaranteed to be found at $K$ steps at maximum.
The $i \geq K$ condition thus allows redefining Eq.~\eqref{eq:maximum_entropy_rate} as}
\begin{equation}
  \hlg{\bar{\mu}^*_{\Delta}(\mathcal{P}_\nu) =  \max_{i = \{ 1, \dots, K \} } \left( \bar{\mu}_{\Delta}\left( \leftscript{i}{\hat{\mathcal{P}}} \right) \right),}
  \label{eq:max_entropy_rate_K}
  \vspace{-2mm}
\end{equation}
\hlg{where $\leftscript{i}{\hat{\mathcal{P}}}$ is the sampled point-set at iteration $i$.}
After terminating the routine at iteration $i \geq K$, the sampled points $\hat{\mathcal{P}} = \leftscript{i}{\hat{\mathcal{P}}}$ equal to all the points sampled up to iteration $i$.

This entropy-maximizing approach normalizes the spectrum of $\Delta$, and thus introduces a certain level of redundancy defined in \autoref{sec:redundancy_in_a_point_set} (by including points with low $\Delta$).
A certain level of redundancy helps in maintaining the original spatial distribution of the points (similar to voxelization), which is beneficial in reducing the effects of noise and outliers.
It has been verified experimentally that the entropy maximization of \acs{gfh} is more resilient than maximizing the cumulative sum of \acs{gfh}, which tends to under-constrain the problem and is sensitive to noise and outliers.




\begin{algorithm}
  \scriptsize
  \algdef{SE}[SUBALG]{Indent}{EndIndent}{}{\algorithmicend\ }%
  \algtext*{Indent}
  \algtext*{EndIndent}

  \algnewcommand\AND{\textbf{and}~}
  \algnewcommand\Not{\textbf{not}~}
  \algnewcommand\Or{\textbf{or}~}
  \algnewcommand\Input{\State{\textbf{Input:~}}}%
  \algnewcommand\Output{\State{\textbf{Output:~}}}%
  \algnewcommand\Parameters{\State{\textbf{Parameters:~}}}%
  \algnewcommand\Begin{\State\textbf{Begin:~}}%
  \algnewcommand{\LineComment}[1]{\State \(\triangleright\) #1}
  \algnewcommand{\LineCommentB}[1]{\State \(\blacktriangleright\) #1}

    \newcommand{\func}[2]{\text{#1}\left(#2\right)}

  \caption{Information-maximizing point selection}\label{alg:entropy_maximization}

  \begin{algorithmic}[1]

    \Input
    \Indent

    \State $\set{P} = \left\{ \vect{p} \right\},\, \vect{p} \in \mathbb{R}^3 $
    \Comment{input point set}

    \State $\nu \in \mathbb{R}^+ $
    \Comment{voxel size in meters}

    \State $K \in \mathbb{Z}^+ $
    \Comment{number of histogram bins}

    \State $\lambda_{\bar{H}} \in \left< 0, 1 \right> $
    \Comment{entropy-rate termination criteria (Eq.~\eqref{eq:maximizing_entropy})}

    \State $C, R \in \mathbb{Z}^+ $
    \Comment{number of columns and rows (if $\set{P}$ in matrix form)}

    \State $\theta_v \in \mathbb{R}^+ $
    \Comment{vertical field of view of the sensor (if $\set{P}$ in matrix form)}

    \EndIndent

    \Output
    \Indent

    \State $\hat{\set{P}} \subseteq \set{P}$
    \Comment{point subset maximizing \acs{gfh} entropy, Eq.~\eqref{eq:maximizing_entropy}}

    \EndIndent

    \Begin
    \Indent


    \State $\set{P}_\nu = \func{voxelize}{\set{P}, \nu}$




    \State $\mathbb{K}_\nu = \func{KDTree}{\set{P}_\nu} $
    \Comment{construct KD-tree for efficient NN search}

    \State $\Delta_{\set{P}_\nu} = \func{GFH}{\set{P}_\nu, \mathbb{K}_\nu, \hlg{C, R, \theta_\nu}} $
    \Comment{Eq.~\eqref{eq:gh_neighborhood}-\eqref{eq:neighborhood_radius}}








    \State $\bar{\Delta}_{\set{P}_\nu} = \func{normalizeGFH}{\Delta_{\set{P}_\nu}} $
    \Comment{Eq.~\eqref{eq:gfh_normalized_norm}}

    \LineCommentB{Construct a histogram of \acs{gfh} values}

    \State $ \set{H}_{\Delta} = \func{histogram}{\bar{\Delta}_{\set{P}_\nu}, K}$
    \Comment{discretize $\bar{\Delta}_{\set{P}_\nu}$ into $K$ fixed-sized bins}

    \State $ \hat{\set{H}}_{\Delta} = \func{histogram}{\emptyset, K}$
    \Comment{empty histogram of $K$ fixed-sized bins}

    \LineCommentB{Compute maximum entropy rate $\bar{\mu}_\Delta^*$}
    \State $ \hlg{ \bar{\mu}^*_{\Delta} = 0 }$

    \For{each $ k \in \left< 1, K \right>$}
    \Comment{iterate each bin exactly once}

      \State $ \leftscript{k}{\set{H}_{\Delta}} = \func{sort}{\leftscript{k}{\set{H}_{\Delta}}}$
      \Comment{sort bin $k$ in desc. order by Eq.~\eqref{eq:sorting_keys}}




      \State $ \leftscript{k}{\hat{\set{H}}_\Delta} = \leftscript{k}{\hat{\set{H}}_\Delta} \cup \left\{ \leftscript{k}{\set{H}_\Delta}\func{.pop}{} \right\} $
      \Comment{move \hlg{highest-value point between bins $k$}}


      \State $ \hlg{ \bar{\mu}^*_{\Delta} = \text{max}\left\{ \bar{\mu}^*_{\Delta},\;\bar{\mu}_{\Delta} \left( \hat{\set{H}}_\Delta \right) \right\}} $
      \Comment{ \hlg{ Eq.~\eqref{eq:max_entropy_rate_K} }}


    \EndFor


    \LineCommentB{Entropy-maximizing selection}
    \State $k = K$
    \Comment{current bin-lookup index}

    \While{$ |\set{H}_\Delta| > 0 $ \AND $ \hlg{ \bar{\mu}_{\Delta} \left( \hat{H}_\Delta \right) / \bar{\mu}_{\Delta}^* > \lambda_{\bar{H}} } $ }
    \Comment{\hlg{terminating via Eq.~\eqref{eq:maximizing_entropy} and \eqref{eq:relative_entropy_rate}}}





      \State $ \leftscript{k}{\hat{\set{H}}_\Delta} = \leftscript{k}{\hat{\set{H}}_\Delta} \cup \left\{ \leftscript{k}{\set{H}_\Delta}\func{.pop}{} \right\} $
      \Comment{move first point in bin $k$}


      \State $ k = k - 1 $ \textbf{if} $k > 1$ \textbf{else} K
      \Comment{cyclic right-left iteration}

    \EndWhile

    

    \State $ \hat{\set{P}} = \bigcup_{k \in \left< 1, K \right>} \leftscript{k}{\hat{\set{H}}_\Delta} $
    \Comment{extract all selected points}


    \EndIndent

  \end{algorithmic}

\end{algorithm}






\vspace{-5mm}
\section{Experimental Analyses}\label{sec:results}

Let us compare the proposed approach with \hlr{three} state-of-the-art point cloud sampling methods:
\begin{itemize}
  \item \textbf{V}$\bullet$: uniform sampling\footnote{Open-source implementation taken from KISS-ICP~\cite{vizzo2023KISSICPDefensePointtoPoint}.} with voxel size $\nu = \bullet\,\,\si{\centi\meter}$,
  \item \textbf{NS}$\bullet$: normal-space voxelization~\cite{rusinkiewicz2001EfficientVariantsICP} with angular resolution $\pi = \bullet\,\,\si{\degree}$ in both azimuth and elevation, and
  \item \hlr{\textbf{CovS}$\bullet$: covariance-based sampling\footnote{\hlr{Open-source implementation taken from PointMatcher~\cite{pomerleau2013ComparingICPVariants}.}}~\cite{gelfand2003GeometricallyStableSampling} with sampled-to-all point ratio of $\rho = \bullet\,\,\si{\percent}$}.
\end{itemize}

All the state-of-the-art methods and the proposed approach were integrated into two state-of-the-art odometry (no loop closures) pipelines: KISS-ICP~\cite{vizzo2023KISSICPDefensePointtoPoint} and \acs{loam}~\cite{zhang2014LOAMLidarOdometry}.
KISS-ICP is a state-of-the-art implementation of the \acs{icp} algorithm, a typical case of a \textit{dense} method utilizing the \textit{point-to-point} metric.
\acs{loam} is a \textit{feature}-based odometry method extracting plane and line features.
\acs{loam} represents a basis for the majority of the \textit{feature}-based state-of-the-art methods.
Since the proposed sampling method is algorithm-independent, it has the potential for positively improving all the other \ac{lidar}-based odometry and \acs{slam} methods building upon \acs{icp} and \acs{loam} algorithms.
To remain close to the core principles and to reduce the effects of any additional concepts, these two representative odometry pipelines have been chosen for their minimalism on purpose.

To ensure a fair comparison, the best parametrizations balancing convergence and real-time processing were fine-tuned manually for \hlr{all} methods, both odometry pipelines, and all datasets.
These parametrizations are given in~\autoref{tab:quantitative_analysis_params}.
All the experiments were performed on AMD Ryzen 7 PRO 4750U \hlr{(comparable performance verified on Intel\textsuperscript{\textregistered} Core i7-10710U)}.


\vspace{-2mm}
\subsection{Datasets}\label{sec:datasets}

The datasets used in evaluation are summarized in Tab.~\ref{tab:datasets}.
Their selection includes custom data from \acsp{mav} (D1-D3) covering full six-\acs{dof} movements in different degraded contexts, and KITTI (D4) and Hilti-Oxford (D5) as two of the most prevalent datasets used in evaluating \acs{lidar}-based methods in the related literature.
Only 3D \acs{lidar} data are used.

\begin{table}[htb]
  \vspace{-1mm}
  \centering

  \caption{Table of used datasets.}
  \label{tab:datasets}

  \def\arraystretch{0.8}
  \setlength\tabcolsep{1.5mm}
  \vspace{-1mm}
  \scriptsize
  \begin{tabular}{llllcr}
    \toprule
    \textbf{ID} & \textbf{Dataset} & \textbf{Work} & \textbf{Platform} & \textbf{Real world} & \textbf{Point count}\\
    \midrule
    D1 & X-ICP        & \cite{tuna2024XICPLocalizabilityAwareLiDAR} & Drone    & \xmark      & $64\!\times\!1024$ @ \SI{10}{\hertz}  \\
    D2 & Dronument    & \cite{petracek2023NewEraCultural}\footnote{} & Drone    & \greencheck & $16\!\times\!1024$ @ \SI{10}{\hertz}  \\
    D3 & DARPA SubT   & \cite{petrlik2023UAVsSurfaceCooperative}\footnotemark[\value{footnote}] & Drone    & \greencheck & $64\!\times\!512$  @ \SI{10}{\hertz}  \\
    D4 & KITTI        & \cite{geiger2012AreWeReady} & Car      & \greencheck & $16\!\times\!1024$ @ \SI{10}{\hertz}  \\
    D5 & Hilti-Oxford & \cite{zhang2023HiltiOxfordDatasetMillimeterAccurate} & Handheld & \greencheck & $32\!\times\!2000$ @ \SI{10}{\hertz}                \\
    \bottomrule
  \end{tabular}
  \vspace{-3mm}
\end{table}
\footnotetext{Dataset available at \href{https://github.com/ctu-mrs/slam_datasets}{github.com/ctu-mrs/slam\_datasets}.}



\vspace{-3mm}
\subsection{Parametrization of RMS}\label{sec:parametrization}


It has been validated empirically that out of the three sensor-agnostic parameters in \autoref{alg:entropy_maximization}, $K$ and $\nu$ have limited effect on the performance.
Thus, $K = 10$ remains fixed in all the presented experiments and $\nu$ is selected such that uniform sampling \textbf{V}$\bullet$ is the most accurate and computes in real time in the given dataset.
\autoref{tab:parametrization} presents an ablation study on the maximum relative entropy rate $\lambda_{\bar{H}}$.
The table demonstrates that the algorithm is stable once $\lambda_{\bar{H}}$ lies in a reasonable interval, here \SIrange[range-phrase=-, range-units=single]{0.2}{0.7}{\percent}.
Based on \autoref{tab:parametrization}, we use $\lambda_{\bar{H}} = \SI{0.4}{\percent}$ in all our KISS-ICP~\cite{vizzo2023KISSICPDefensePointtoPoint} experiments as a balance between runtime, accuracy, and stability.
Since the stable interval is pipeline-dependent, similar grid-search has been done to find optimal $\lambda_{\bar{H}}$ for the \textit{feature}-based LOAM~\cite{zhang2014LOAMLidarOdometry} estimation pipeline used in \autoref{sec:quantitative_analysis}.
In LOAM, one instance of~\autoref{alg:entropy_maximization} runs independently for each of the feature types, with $\lambda_{\bar{H}}$ being fixed to $\lambda_{\bar{H}} = \SI{0.8}{\percent}$ for plane and $\lambda_{\bar{H}} = \SI{15}{\percent}$ for line features.


\begin{table}[htb]
  \vspace{-1mm}
  \centering

  \caption{Influence of the maximum relative entropy rate $\lambda_{\bar{H}}$ on performance of the proposed method in experiment presented in \autoref{fig:convergence_analysis}.}
  \label{tab:parametrization}

  \def\arraystretch{0.8}
  \vspace{-1mm}
  \begin{tabular}{l r r r r r r r r}
    \toprule
    $\lambda_{\bar{H}} (\si{\percent})$ & 0.1 & 0.2 & 0.3 & 0.4 & 0.5 & 0.7 & 1.0 \\
    \midrule
    RMSE (\si{\meter})                  & 0.31 & \textbf{0.22} & 0.25 & 0.24 & 0.27 & 0.29 & 0.43 \\
    avg. time (\si{\milli\second})      & 42 & 37 & 32 & 29 & 27 & 24 & 23\\
    compr. rate (\si{\percent})         & 95.3 & 97.0 & 97.8 & 98.2 & 98.5 & 98.9 & 99.2\\
    \bottomrule
  \end{tabular}
  \vspace{-3mm}
\end{table}



\vspace{-3mm}
\subsection{Convergence Analysis}\label{sec:convergence_analysis}


\autoref{fig:convergence_analysis} demonstrates an experiment designed to compare performance of the \hlr{four} sampling techniques \hlr{(all fine-tuned to the environment)}.
In the experiment, a \acs{mav} performs a loop inside a challenging simulation world (D1) designed to contain various geometrical degeneracies (translational along narrow corridors and rotational within a circular room).
The experiment shows that the proposed method outperforms \hlr{the baseline} methods in terms of speed, accuracy, and robustness \hlr{(even to geometrical degeneracies)}, all while sampling the least amount of points.
The \hlr{data show} superior timing and compression consistency of the proposed method, with both reaching almost constant values with \hlr{a} limited number of \hlr{outliers.}
This is \hlr{a} particularly important attribute for deployment of small and agile resource-constrained robots with real-time constraints, such as \acsp{mav}.
\autoref{fig:convergence_analysis_performance} shows that information rate \hlr{(measured as eigenvalue per point sampled on input) extracted in the optimization from the problem Hessian is highest in RMS (only z-axis rotational eigenvalue $R_z$ is shown).
Additionally, \autoref{fig:gfh_example} showcases a single-frame sampling of \autoref{alg:entropy_maximization}.}

\ifdefined\compilewithallfigures

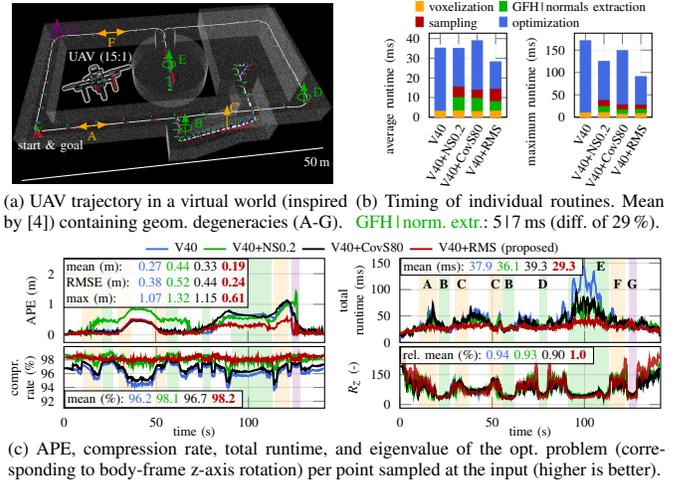
\begin{figure}
  \begin{minipage}[t]{0.48\columnwidth}
  \subfloat[\scriptsize\hlr{\acs{mav} trajectory in a virtual world (inspired by~\cite{tuna2024XICPLocalizabilityAwareLiDAR}) containing geom. degeneracies (A-G).}
            \label{fig:convergence_analysis_traj}]
            {\hspace{-0.5mm}\vspace{1.0mm}\input{./fig/convergence_analysis/traj.tex}}
  \end{minipage}%
  \hspace{3.5mm}
  \begin{minipage}[t]{0.50\columnwidth}
    \subfloat[\scriptsize\hlr{Timing of individual routines. Mean \textcolor{color_green}{\acs{gfh}}\textcolor{color_green}{\,|\,}\textcolor{color_green}{norm. extr.}: 5\,|\,\SI{7}{\milli\second} (diff. of \SI{29}{\percent}).}
            \label{fig:convergence_analysis_runtime}]
            {\hspace{2mm}\vspace{-3.5mm}\input{./fig/convergence_analysis/runtime.tex}}
  \end{minipage}\\
  \begin{minipage}[t]{1.0\columnwidth}
  \centering
  \vspace{-0.3mm}
  \subfloat[\scriptsize\hlr{APE, compression rate, total runtime, and eigenvalue of the opt. problem (corresponding to body-frame z-axis rotation) per point sampled at the input (higher is better).}
            \label{fig:convergence_analysis_performance}]
            {\hspace{-2.5mm}\vspace{-3mm}\input{./fig/convergence_analysis/performance.tex}}
  \end{minipage}
  \vspace{-0.5mm}
  \caption{\footnotesize\hlr{%
    Output of KISS-ICP~\cite{vizzo2023KISSICPDefensePointtoPoint} 6-\acs{dof} odometry when preceded by different point cloud sampling methods.
    Parametrization: best performing for each method, robot: multirotor \acs{mav}, sensor range: \SI{30}{\meter}, sensor noise: none.
    (a) Shows areas of translational (A, C, F) and body-frame z-axis rotational (B, D, E, G) degeneracy.
        At D and G, the degeneracy arises (see low values of the $R_z$ eigenvalue) from large \acs{mav} tilt, which orients the \acs{lidar} such that its data are degenerate around the z-axis.
        At G, a "loop closing" emerges naturally (see APE).
    (b) Due to the high compression rate and by balancing the translational space, RMS samples points such that they yield the fastest optimization convergence.
    (c) RMS yields the lowest drift, removes the largest amount of points, produces stable and lowest runtime, and preserves the highest information rate for optimization (only $R_z$ shown).
    \label{fig:convergence_analysis}}%
  }%
\vspace{-1mm}
\end{figure}

\fi
\ifdefined\compilewithallfigures

\begin{figure}[t]
  \centering
  \input{./fig/gfh/gfh_analysis.tex}
  \vspace{-2mm}
  \caption{\hlg{Rel. entropy}, rel. entropy rate, and rel. redundancy of \ac{gfh} at time \SI{60}{\second} of experiment in \autoref{fig:convergence_analysis}.
           \textbf{V40+RMS} sampled about $\SI{2}{\percent}$ of points out of 65k total.
           Dashed lines represent the non-sampled points.}
  \label{fig:gfh_example}
  \vspace{-4mm}
\end{figure}
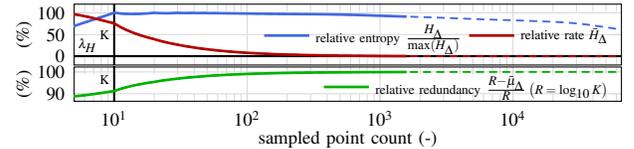

\fi


\vspace{-2mm}
\subsection{Quantitative analysis}\label{sec:quantitative_analysis}

\autoref{tab:quantitative_analysis} presents a quantitative analysis comparing the effects of the sampling methods on the two odometry pipelines.
Together with~\autoref{tab:quantitative_analysis_params}, the two tables show that the fixed parametrization adapts well to various different sensors, environments, and conditions.
This is a significant practical advantage, which improves the method's applicability by reducing the need for tuning the proposed method to every domain.
The data show superior performance of the proposed method in
\begin{itemize}
    \item improving performance in well-conditioned settings,
    \item reducing odometry drift in degenerated conditions,
    \item sampling the least amount of points in general, and
    \item computing the fastest while being the most accurate.
\end{itemize}
\hlr{The reported timings are for the entire pipeline, including point or feature sampling, and the optimization.
The accuracy gains are associated with lower comp. time enabling use of all data in real-time as well as with high noise and outlier removal.}

\begin{table}[t]
  \centering
  \def\arraystretch{0.8} 
  \setlength\tabcolsep{1.4mm}
  \captionsetup{font=footnotesize}
  \vspace{1.5mm}
  \caption{Quantitative performance of KISS-ICP and LOAM pose estimation pipelines when preceded by \hlr{four} different 3D \acs{lidar} point cloud sampling techniques:
           uniform sampling \textbf{V}$\bullet$,
           normal-space sampling \textbf{V}$\bullet$\textbf{+NS}$\bullet$~\cite{rusinkiewicz2001EfficientVariantsICP},
           \hlr{covariance sampling \textbf{V}$\bullet$\textbf{+CovS}$\bullet$~\cite{gelfand2003GeometricallyStableSampling}, and}
           \hlr{redundancy-minimizing sampling} \textbf{V}$\bullet$\textbf{+RMS} (proposed).
           \hlr{Metrics: APE $|\delta|\;(\si{\meter})$, RPE ${^\Delta}\!\delta\;(\si{\meter})$, total runtime $\tau\;(\si{\milli\second})$, and compression rate $\chi\;(\si{\percent})$.}
           \hlr{The best} results are in \textbf{bold}.
           Trajectories of experiments \textbf{D2} and \textbf{D4: KITTI \#00} are shown in \autoref{fig:motivation}.}
  \label{tab:quantitative_analysis}
  \vspace{-1mm}
  \scriptsize
  \begin{tabular}{ll@{\hskip 3mm}rrrr@{\hskip 7mm}rrrr}
   \toprule
    \multirow{2}{*}[-0mm]{\rotatebox{90}{\textbf{Dataset}}}
    & \multicolumn{1}{c}{\multirow{2}{*}[-0.5mm]{\rotatebox{90}{\textbf{Metric}}}} 
    & \multicolumn{4}{c}{\textbf{KISS-ICP}} & \multicolumn{4}{c}{\textbf{LOAM}}\\\cmidrule(lr){3-6} \cmidrule(lr){7-10}
    & & \multirow{2}{*}{\textbf{V}$\bullet$} &
          \multirow{2}{*}{\shortstack[c]{\textbf{V40}\textbf{+}\\\textbf{NS$\bullet$}}} &
          \multirow{2}{*}{\shortstack[c]{\hlr{\textbf{V40}\textbf{+}}\\\hlr{\textbf{CovS}$\bullet$}}} &
          \multirow{2}{*}{\shortstack[c]{\textbf{V40}\textbf{+}\\\textbf{RMS}}} &
          \multirow{2}{*}{\textbf{V}$\bullet$} &
          \multirow{2}{*}{\shortstack[c]{\textbf{V}$\bullet$\textbf{+}\\\textbf{NS}$\bullet$}} &
          \multirow{2}{*}{\shortstack[c]{\hlr{\textbf{V}$\bullet$\textbf{+}}\\\hlr{\textbf{CovS}$\bullet$}}} &
          \multirow{2}{*}{\shortstack[c]{\textbf{V}$\bullet$\textbf{+}\\\textbf{RMS}}} \\
    & & & & & & & \\
    \midrule
    \multirow{9}{*}[-1mm]{\rotatebox{90}{\textbf{D2: Star\'{a} Voda \#00}}} 
    & $|\delta|_{rmse}$ &  
    \multicolumn{4}{c}{\multirow{9}{*}[-2mm]{\rotatebox{0}{\shortstack[c]{failed in estimating\\vertical motion}}}}
    & 0.85 & 0.31 & 0.19 & \textbf{0.12} \\
    & $|\delta|_{mean}$ & & & & & 0.39 & 0.18 & 0.15 & \textbf{0.09} \\
    & $|\delta|_{max}$  & & & & & 2.56 & 0.91 & 0.45 & \textbf{0.39}%
    \vspace{0.2mm}
    \\\rule{0pt}{2.0mm}
    & ${^\Delta}\!\delta_{rmse}$ & & & & & 0.06 & 0.02 & \textbf{0.01} & 0.02 \\
    & ${^\Delta}\!\delta_{mean}$ & & & & & \textbf{0.01} & \textbf{0.01} & \textbf{0.01} & \textbf{0.01} \\
    & ${^\Delta}\!\delta_{max}$  & & & & & 1.57 & 0.39 & \textbf{0.12} & 0.18%
    \vspace{0.2mm}
    \\\rule{0pt}{2.0mm}
    & $\tau_{mean}$ & & & & & 69.0  & 61.4  & 37.9 & \textbf{31.0} \\
    & $\tau_{max}$  & & & & & 109.6 & 138.8 & 74.3 & \textbf{69.7}%
    \vspace{0.2mm}
    \\\rule{0pt}{2.0mm}
    & $\chi_{mean}$ & & & & & 74.8  & 82.0  & 90.9 & \textbf{95.7} \\
    \midrule
    \multirow{9}{*}[-1mm]{\rotatebox{90}{\textbf{D3: urban corridor}}} 
    & $|\delta|_{rmse}$ & 0.59 & 0.63 & 0.67 & \textbf{0.42} & 1.20 & 2.01 & \textbf{0.75} & 1.20 \\
    & $|\delta|_{mean}$ & 0.50 & 0.52 & 0.63 & \textbf{0.38} & 1.14 & 1.78 & \textbf{0.69} & 1.16 \\
    & $|\delta|_{max}$  & 1.24 & 1.34 & 1.11 & \textbf{0.86} & 1.84 & 4.00 & \textbf{1.41} & 1.78%
    \vspace{0.2mm}
    \\\rule{0pt}{2.0mm}
    & ${^\Delta}\!\delta_{rmse}$ & \textbf{0.04} & \textbf{0.04} & \textbf{0.04} & \textbf{0.04} & \textbf{0.03} & 0.04 & \textbf{0.03} & \textbf{0.03} \\
    & ${^\Delta}\!\delta_{mean}$ & \textbf{0.03} & \textbf{0.03} & 0.04          & \textbf{0.03} & \textbf{0.02} & 0.03 & 0.03 & \textbf{0.02} \\
    & ${^\Delta}\!\delta_{max}$  & \textbf{0.23} & 0.29          & 0.24          & \textbf{0.23} & \textbf{0.12} & 0.18 & 0.17 & \textbf{0.12}%
    \vspace{0.2mm}
    \\\rule{0pt}{2.0mm}
    & $\tau_{mean}$ & \textbf{13.5} & 18.1 & 24.1 & 16.7 & 13.0 & 15.1 & \textbf{11.0} & 13.0\\
    & $\tau_{max}$  & 61.3          & 56.8 & 77.9 & \textbf{55.7} & 36.3 & 32.7 & \textbf{24.9} & 28.6%
    \vspace{0.2mm}
    \\\rule{0pt}{2.0mm}
    & $\chi_{mean}$ & 95.8 & \textbf{96.2} & 93.4 & 95.9 & 94.8 & 99.0 & \textbf{99.1} & 98.5\\
    \midrule
    \multirow{9}{*}[-1.5mm]{\rotatebox{90}{\textbf{D4: KITTI \#00}}}
    & $|\delta|_{rmse}$ & 248.60 & 14.31 & 56.01  & \textbf{8.35}   & 20.72 & 25.11 & \textbf{12.28} & 12.76\\
    & $|\delta|_{mean}$ & 205.97 & 11.27 & 45.29  & \textbf{7.48}   & 16.34 & 21.05 & \textbf{9.50} & 9.78 \\
    & $|\delta|_{max}$  & 458.96 & 32.59 & 116.34 & \textbf{16.20}  & 49.87 & 51.82 & 29.96 & \textbf{29.45}%
    \vspace{0.2mm}
    \\\rule{0pt}{2.0mm}
    & ${^\Delta}\!\delta_{rmse}$ & 1.41 & \textbf{1.27} & \textbf{1.27}  & \textbf{1.27}  & \textbf{1.27} & \textbf{1.27} & \textbf{1.27}  & \textbf{1.27} \\
    & ${^\Delta}\!\delta_{mean}$ & 1.26 & \textbf{1.17} & \textbf{1.17}  & \textbf{1.17}  & \textbf{1.17} & \textbf{1.17} & \textbf{1.17}  & \textbf{1.17} \\
    & ${^\Delta}\!\delta_{max}$  & 14.86 & 14.71        & \textbf{14.64} & 14.67          & 14.79         & 14.72         & 14.70  & \textbf{14.69}%
    \vspace{0.2mm}
    \\\rule{0pt}{2.0mm}
    & $\tau_{mean}$ & 43.3   & 57.1 & \textbf{34.2} & 35.7          & 83.0  & 76.3  & 74.9  & \textbf{66.9}  \\
    & $\tau_{max}$  & 769.1 & 270.8 & 88.4          & \textbf{87.7} & 180.8 & 196.9 & 168.1 & \textbf{159.6} \\
    & $\chi_{mean}$ & 96.6   & 97.2 & 99.2          & \textbf{99.3} & 89.4  & 99.2  & 99.1  & \textbf{99.4}   \\
    \midrule
    \multirow{9}{*}[-1.5mm]{\rotatebox{90}{\textbf{D4: KITTI \#09}}} 
    & $|\delta|_{rmse}$ & 481.02 & 17.48 & 25.66 & \textbf{15.76} & 16.08 & 21.15 & 12.64 & \textbf{10.77} \\
    & $|\delta|_{mean}$ & 418.98 & 13.89 & 19.31 & \textbf{13.12} & 11.94 & 15.72 & 9.18  & \textbf{8.07}  \\
    & $|\delta|_{max}$  & 772.04 & 46.24 & 68.67 & \textbf{32.61} & 38.86 & 50.01 & 31.49 & \textbf{25.59}%
    \vspace{0.2mm}
    \\\rule{0pt}{2.0mm}
    & ${^\Delta}\!\delta_{rmse}$ & 2.19 & 1.65  & 1.75 & \textbf{1.58} & \textbf{1.58} & \textbf{1.58} & \textbf{1.58} & \textbf{1.58} \\
    & ${^\Delta}\!\delta_{mean}$ & 1.71 & 1.57  & 1.62 & \textbf{1.52} & \textbf{1.52} & \textbf{1.52} & \textbf{1.52} & \textbf{1.52} \\
    & ${^\Delta}\!\delta_{max}$  & 25.22 & 4.28 & 7.72 & \textbf{3.58} & \textbf{3.60} & \textbf{3.60} & 3.62 & 3.65%
    \vspace{0.2mm}
    \\\rule{0pt}{2.0mm}
    & $\tau_{mean}$ & 71.3   & 67.0  & 59.8  & \textbf{42.8}  & 70.2  & 59.1  & 59.33 & \textbf{50.8}\\
    & $\tau_{max}$  & 1748.3 & 811.3 & 700.5 & \textbf{118.0} & 143.4 & 119.2 & 127.3 & \textbf{92.7}%
    \vspace{0.2mm}
    \\\rule{0pt}{2.0mm}
    & $\chi_{mean}$ & 95.9   & 96.4 & 97.1 & \textbf{98.8} & 87.1 & 99.0 & 98.5 & \textbf{99.4} \\
    \midrule
    \multirow{9}{*}[-0.5mm]{\rotatebox{90}{\textbf{D5: Hilti-Oxford \#04}}} 
    & $|\delta|_{rmse}$ &
    \multicolumn{4}{c}{\multirow{9}{*}[-2mm]{\rotatebox{0}{\shortstack[c]{failed in estimating\\quick rotational motions}}}}
    & 0.34 & 0.63 & 0.29 & \textbf{0.23} \\
    & $|\delta|_{mean}$ & & & & & 0.30 & 0.59 & 0.26 & \textbf{0.21} \\
    & $|\delta|_{max}$  & & & & & 0.74 & 1.08 & 0.67 & \textbf{0.58}%
    \vspace{0.2mm}
    \\\rule{0pt}{2.0mm}
    & ${^\Delta}\!\delta_{rmse}$ & & & & & \textbf{0.14} & \textbf{0.14} & \textbf{0.14} & \textbf{0.14} \\
    & ${^\Delta}\!\delta_{mean}$ & & & & & \textbf{0.10} & \textbf{0.10} & \textbf{0.10} & \textbf{0.10} \\
    & ${^\Delta}\!\delta_{max}$  & & & & & 0.56          & 0.61          & 0.54 & \textbf{0.51}%
    \vspace{0.2mm}
    \\\rule{0pt}{2.0mm}
    & $\tau_{mean}$ & & & & & 67.9  & 55.7  & 54.1 & \textbf{42.8} \\
    & $\tau_{max}$  & & & & & 138.5 & 133.9 & 120.0 & \textbf{77.1}%
    \vspace{0.2mm}
    \\\rule{0pt}{2.0mm}
    & $\chi_{mean}$ & & & & & 90.5 & 95.4 & 95.3 & \textbf{97.7} \\
    \midrule
    \multirow{9}{*}[-0.5mm]{\rotatebox{90}{\textbf{D5: Hilti-Oxford \#14}}} 
    & $|\delta|_{rmse}$ &
    \multicolumn{4}{c}{\multirow{9}{*}[-2mm]{\rotatebox{0}{\shortstack[c]{failed in estimating\\quick rotational motions}}}}
    & 2.22 & 2.85 & 0.84 & \textbf{0.76} \\
    & $|\delta|_{mean}$ & & & & & 1.86 & 2.35 & \textbf{0.62} &\textbf{0.62}  \\
    & $|\delta|_{max}$  & & & & & 3.69 & 4.82 & 2.06 &\textbf{1.64}%
    \vspace{0.2mm}
    \\\rule{0pt}{2.0mm}
    & ${^\Delta}\!\delta_{rmse}$ & & & & & \textbf{0.16} & \textbf{0.16} & \textbf{0.16} & \textbf{0.16} \\
    & ${^\Delta}\!\delta_{mean}$ & & & & & \textbf{0.09} & \textbf{0.09} & \textbf{0.09} & \textbf{0.09} \\
    & ${^\Delta}\!\delta_{max}$  & & & & & \textbf{1.91} & \textbf{1.91} & \textbf{1.91} & \textbf{1.91}%
    \vspace{0.2mm}
    \\\rule{0pt}{2.0mm}
    & $\tau_{mean}$ & & & & & 20.7 & 18.2 & 17.3 & \textbf{17.2} \\
    & $\tau_{max}$  & & & & & 45.1 & 47.6 & 40.4 & \textbf{30.3}%
    \vspace{0.2mm}
    \\\rule{0pt}{2.0mm}
    & $\chi_{mean}$ & & & & & 97.9 & 98.7 & 98.7 & \textbf{98.8} \\
    \bottomrule
  \end{tabular}
  \vspace{-3mm}
\end{table}

\begin{table}[htb]
  \centering
  \def\arraystretch{0.8}
  \setlength\tabcolsep{1.0mm}
  \vspace{1.5mm}
  \captionsetup{font=footnotesize}
  \caption{\hlr{Table of parameters used in the experiments in~\autoref{tab:quantitative_analysis}.
           The values marked with $^*$ are for the \textbf{V}$\bullet$ method only (the \textbf{V40} parametrizations use $\nu = \SI{40}{\centi\meter}$).
           Among the compared methods, only the proposed method's parametrization is fixed $\left(\lambda_{\bar{H}}\right)$, showing the method's unique adaptability to different sensors and environments.}}
  \label{tab:quantitative_analysis_params}
  \vspace{-1mm}
  \scriptsize
  \begin{tabular}{lrrrr @{\hskip 4mm} rrrr @{\hskip 2mm} rrrr}
   \toprule
   \textbf{Pipeline}
    & \multicolumn{4}{c}{\textbf{KISS-ICP}}
    & \multicolumn{8}{c}{\textbf{LOAM}}\\\rule{0pt}{2mm}

   \textbf{Sampling}
    & \multicolumn{4}{c}{3D points}
    & \multicolumn{4}{c}{plane features}
    & \multicolumn{4}{c}{line features}\\\cmidrule(lr{8pt}){2-5} \cmidrule(r){6-9} \cmidrule(lr){10-13}

    \textbf{Parameter}
    & \multicolumn{1}{c}{$\nu$}
    & \multicolumn{1}{c}{$\pi$}
    & \multicolumn{1}{c}{$\rho$}
    & \multicolumn{1}{c}{$\lambda_{\bar{H}}$}
    & \multicolumn{1}{c}{$\nu$}
    & \multicolumn{1}{c}{$\pi$}
    & \multicolumn{1}{c}{$\rho$}
    & \multicolumn{1}{c}{$\lambda_{\bar{H}}$}
    & \multicolumn{1}{c}{$\nu$}
    & \multicolumn{1}{c}{$\pi$}
    & \multicolumn{1}{c}{$\rho$}
    & \multicolumn{1}{c}{$\lambda_{\bar{H}}$}\\\rule{0pt}{2mm}%
    \textbf{Units}
    & \multicolumn{1}{c}{$(\si{\meter})$}
    & \multicolumn{1}{c}{$(\si{\degree})$}
    & \multicolumn{1}{c}{$(\si{\percent})$}
    & \multicolumn{1}{c}{$(\si{\percent})$}
    & \multicolumn{1}{c}{$(\si{\meter})$}
    & \multicolumn{1}{c}{$(\si{\degree})$}
    & \multicolumn{1}{c}{$(\si{\percent})$}
    & \multicolumn{1}{c}{$(\si{\percent})$}
    & \multicolumn{1}{c}{$(\si{\meter})$}
    & \multicolumn{1}{c}{$(\si{\degree})$}
    & \multicolumn{1}{c}{$(\si{\percent})$}
    & \multicolumn{1}{c}{$(\si{\percent})$} \\
    \midrule
    \textbf{D2} & \multicolumn{1}{r}{\xmark\;} & \multicolumn{1}{r}{\xmark\;} & \multicolumn{1}{r}{\xmark\;} & \multicolumn{1}{r}{\xmark\;} & 0.4 & 0.2 & 40 & \multirow{6}{*}{0.8} & 0.2 & 0.2 & 20 & \multirow{6}{*}{\rotatebox[origin=c]{0}{15}}  \\
    \textbf{D3} & 0.4 & 0.2 & 50 & \multicolumn{1}{r}{\multirow{3}{*}{0.4}} & 0.8 & 0.2 & 60 & & 0.4 & 0.2 & 30 & \\
    \textbf{D4: \#00} & 1.0$^*$ & 1.0 & 10 & & 0.4 & 0.1 & 80 & & 0.2 & 0.1 & 20  & \\
    \textbf{D4: \#09} & 1.0$^*$ & 1.0 & 30 & & 1.0 & 0.1 & 50 & & 0.5 & 0.1 & 2.5 & \\
    \textbf{D5: \#04} & \multicolumn{1}{r}{\xmark\;} & \multicolumn{1}{r}{\xmark\;} & \multicolumn{1}{r}{\xmark\;} & \multicolumn{1}{r}{\xmark\;}         & 0.4 & 1.0 & 50 & & 0.2 & 1.0 & 50 & \\
    \textbf{D5: \#14} & \multicolumn{1}{r}{\xmark\;} & \multicolumn{1}{r}{\xmark\;} & \multicolumn{1}{r}{\xmark\;} & \multicolumn{1}{r}{\xmark\;}         & 0.4 & 1.0 & 80 & & 0.2 & 1.0 & 2.5 & \\
    \bottomrule
  \end{tabular}
  \vspace{-3mm}
\end{table}

\vspace{-2mm}
\bibliographystyle{IEEEtran}
\bibliography{main}


\end{document}

%% file: fig/motivation/sampling_frames.tex
\begin{tikzpicture}[font=\footnotesize]

  \node[anchor=south west,inner sep=0] (a) at (0,0) {
      \def\arraystretch{0}%
      \includegraphics[width=1.0\textwidth]{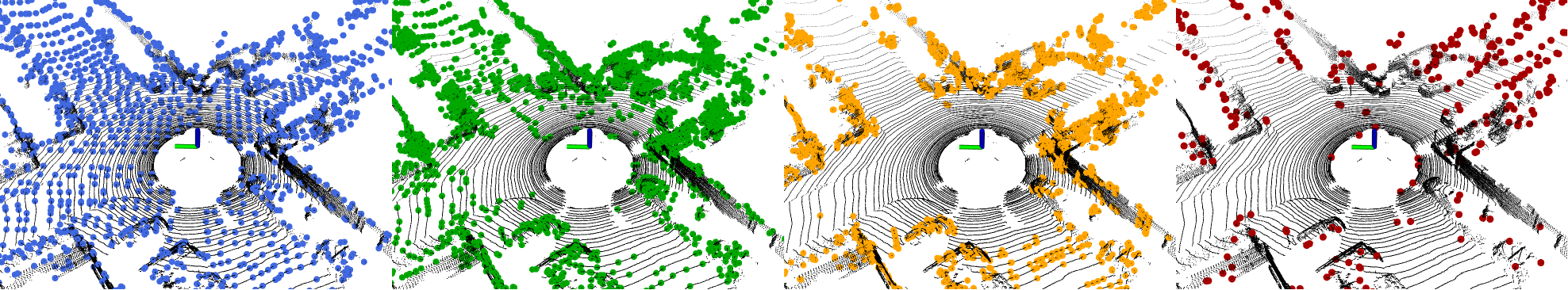}%
    }; 

  \node[fill=white, draw=white, fill opacity=0.9, draw opacity=0.9, text opacity=1.0] at (0.26, 1.385) {\small \textcolor{black}{\textbf{a)}}};



\end{tikzpicture}

%% file: fig/motivation/labels.tex
\begin{tikzpicture}[font=\footnotesize]


  \node[fill=none] at (1.3, 0) {
    \hspace{1.75mm}
    \textcolor{color_blue}{\textbf{Voxelization}}
    \hspace{1.40mm}
    \textcolor{color_green}{\textbf{Rusinkiewicz et al.}}
    \hspace{0.0mm}
    \textcolor{color_orange}{\textbf{Gelfand et al.}}
    \hspace{1.6mm}
    \textcolor{color_red}{\textbf{RMS (proposed)}}
  };

\end{tikzpicture}

%% file: fig/motivation/kitti.tex
\pgfplotsset{
  compat=1.6,
  legend image code/.code={
    \draw[mark repeat=2,mark phase=2]
    plot coordinates {
      (0cm,0cm)
      (0.15cm,0cm)        
      (0.3cm,0cm)         
    };%
  }
}

\begin{tikzpicture}[font=\footnotesize]


  \node[anchor=south west,inner sep=0] (a) at (0,0) {
      \def\arraystretch{0}%
      \includegraphics[width=1.0\textwidth]{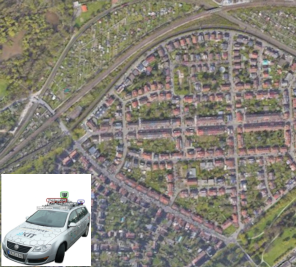}%
    }; 

  \pgfplotstableread[col sep=space]{./fig/motivation/kitti_00/GT.tum}{\tableTRAJGT}
  \pgfplotstableread[col sep=space]{./fig/motivation/kitti_00/NORMAL_SAMPLING.txt}{\tableTRAJNS}
  \pgfplotstableread[col sep=space]{./fig/motivation/kitti_00/UNIFORM_SAMPLING.txt}{\tableTRAJV}
  \pgfplotstableread[col sep=space]{./fig/motivation/kitti_00/COVARIANCE_SAMPLING.txt}{\tableTRAJCOV}
  \pgfplotstableread[col sep=space]{./fig/motivation/kitti_00/RISE.txt}{\tableTRAJRISE}
  
  \begin{axis}[ 
    name=kitti,
    width=1.25\textwidth,
    height=1.16\textwidth,
    xshift=2.0mm,
    yshift=2.5mm,
    enlarge x limits=0.02,
    enlarge y limits=0.02,
    xticklabels={},
    yticklabels={},
    axis line style={latex-latex,draw=white},
    hide axis,
    ]

  \pgfmathsetmacro{\lw}{1.4pt}
  \pgfmathsetmacro{\lwplus}{0.6pt}

  \addplot [color=black, line width=\lw] table [y=y, x=x]{\tableTRAJGT};
  \addplot [color=black, line width=\lw+\lwplus] table [y=y, x=x]{\tableTRAJV};
  \addplot [color=black, line width=\lw+\lwplus] table [y=y, x=x]{\tableTRAJNS};
  \addplot [color=black, line width=\lw+\lwplus] table [y=y, x=x]{\tableTRAJCOV};
  \addplot [color=black, line width=\lw+\lwplus] table [y=y, x=x]{\tableTRAJRISE};

  \addplot [color=black, line width=\lw] table [y=y, x=x]{\tableTRAJGT};
  \addplot [color=color_blue, line width=\lw] table [y=y, x=x]{\tableTRAJV};
  \addplot [color=color_green, line width=\lw] table [y=y, x=x]{\tableTRAJNS};
  \addplot [color=color_orange, line width=\lw] table [y=y, x=x]{\tableTRAJCOV};
  \addplot [color=color_red, line width=\lw] table [y=y, x=x]{\tableTRAJRISE};
    

  \end{axis}

  \draw[draw=black, line width=0.7pt] (0.02, 0.034) rectangle ++(1.46, 1.505);
  \node[fill=white, draw=white, fill opacity=0.5, draw opacity=0.5, text opacity=1.0] at (4.59, 4.13) {\small \textcolor{black}{\textbf{b)}}};

  \draw[->, -latex, color=white, line width=0.7pt] (2.04, 2.705) node[above, color=white, yshift=-1.0mm] {\textbf{(a)}} -- (2.04, 2.2);

  \node[fill=white, draw=black, fill opacity=0.0, draw opacity=0.0, text opacity=1.0,anchor=west] at (-0.00, 1.36)
  {\textcolor{black}{\textbf{KITTI~\cite{geiger2012AreWeReady}}}};

  \draw[|-|, draw=white, line width=1.5pt] (3.6, 0.28) -- node[above, white]{\textbf{\SI{200}{\meter}}} (4.7, 0.28);

\end{tikzpicture}

%% file: fig/motivation/stara_voda.tex
\pgfplotsset{
  compat=1.6,
  legend image code/.code={
    \draw[mark repeat=2,mark phase=2]
    plot coordinates {
      (0cm,0cm)
      (0.15cm,0cm)        
      (0.3cm,0cm)         
    };%
  }
}

\begin{tikzpicture}[font=\footnotesize]

  \node[anchor=south west,inner sep=0] (a) at (kitti.south east) {
      \def\arraystretch{0}%
      \includegraphics[width=1.0\textwidth]{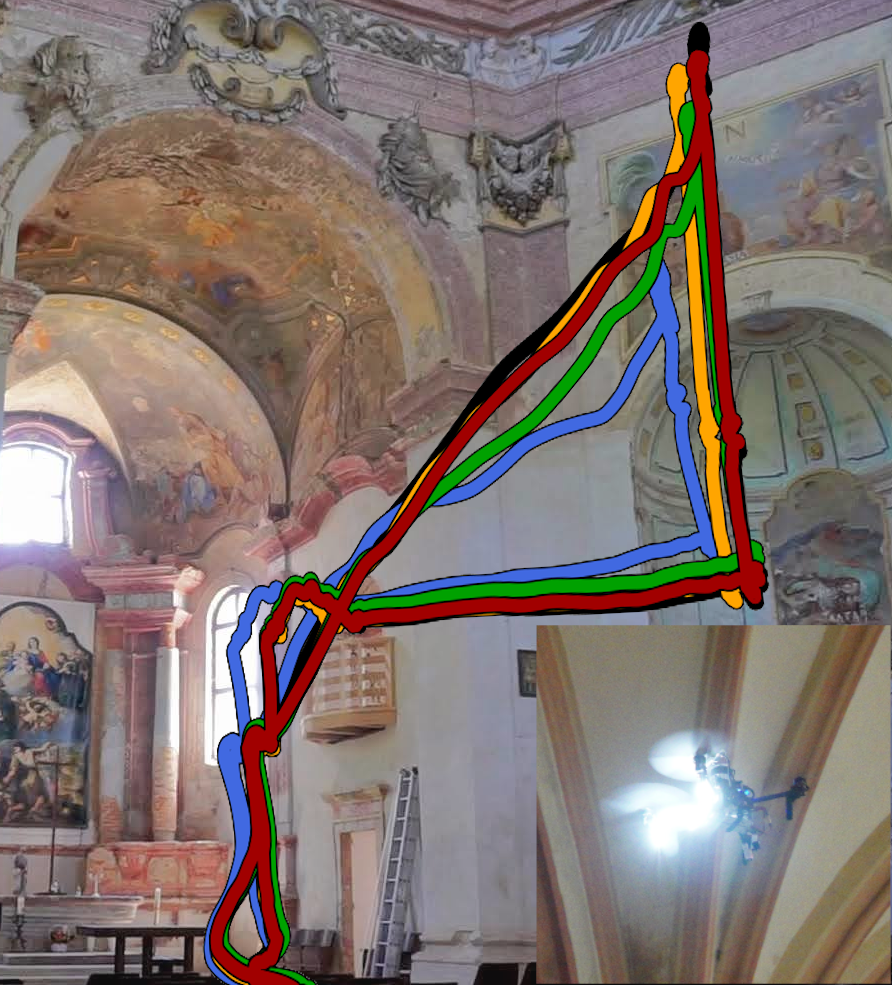}%
    }; 

  \node[fill=white, draw=white, fill opacity=0.5, draw opacity=0.5, text opacity=1.0] at (8.40, 4.365) {\small \textcolor{black}{\textbf{c)}}};
  \draw[draw=white, line width=0.7pt] (7.08, 0.268) rectangle ++(1.5565, 1.59);

  \hypersetup{citecolor=white}
  \node[fill=white, draw=white, fill opacity=0.0, draw opacity=0.0, text opacity=1.0, text=white, anchor=west] at (7.255, 1.63) 
  {\textbf{UAV~\cite{petracek2023NewEraCultural}}};
  \hypersetup{citecolor=black}

  \draw[|-|, color=white, line width=1.5pt] (4.95, 0.35) -- node[right, white, yshift=17mm]{\textbf{\SI{8}{\meter}}} (4.95, 4.55);


\end{tikzpicture}

%% file: fig/architecture.tex
\usetikzlibrary{shapes.geometric,backgrounds,calc,arrows}
\usetikzlibrary{shadows}
\pgfdeclarelayer{background}
\pgfdeclarelayer{foreground}
\pgfsetlayers{background,main,foreground}


\tikzstyle{block} = [%
   rectangle, draw, thick, fill=none, align=center,
   text centered, rounded corners, minimum height=1em,
   execute at begin node={\begin{varwidth}{15em}},
   execute at end node={\end{varwidth}}]

\begin{tikzpicture}[node distance=1.5cm, auto, ->=-latex, font=\tiny]

  \node [block] (cs) {Correspondence\\search};
  \node [block, right of=cs, shift={(1.0, 0.0)}] (res_lin) {Residual computation\\\& linearization};
  \draw [->] (cs.east) -| ++(0.1, 0.2) -- node[above, shift={(0.0, -0.05)}]{$\mathbb{C}_{\set{Q}}^{\set{P}} = \{ \vect{p}_i, \vect{q}_i \}$} ++(0.65, 0) |- (res_lin.west);

  \node [block, right of=res_lin, shift={(0.65, 0.0)}, draw=color_red, fill=color_red!10] (res_samp) {Residual\\sampling};
  \draw [->] (res_lin.east) -| ++(0.1, 0.25) -- node[above, shift={(-0.0, -0.05)}]{$\mathbb{C}_{\set{Q}}^{\set{P}} = \{ \vect{p}_i, \vect{q}_i, \vect{r}_i, \mat{J}_i \}$} ++(0.5, 0) |- (res_samp.west);

  \node [block, right of=res_samp, shift={(0.2, 0.0)}] (opt) {Iterative\\optimization};
  \draw [->] (res_samp.east) -| ++(0.1, 0.2) -- node[above, shift={(0.0, -0.05)}]{$\bar{\mathbb{C}}_{\set{Q}}^{\set{P}} \subseteq \mathbb{C}_{\set{Q}}^{\set{P}}$} ++(0.4, 0) |- (opt.west);

  \node[left of=cs, shift={(0.2, 0.6)}] (label) {\scriptsize \textbf{a)}};

  \node [rectangle, left of=cs, shift = {(-0.0, 0.11)}, node distance=1.3cm, fill opacity=0.0, text opacity=0.0] (k) {};
  \draw [->] (k.east) -- node[left, pos=0.1]{$\set{P}$} (cs.171);

  \node [rectangle, left of=cs, shift = {(-0.0, -0.11)}, node distance=1.3cm, fill opacity=0.0, text opacity=0.0] (k) {};
  \draw [->] (k.east) -- node[left, pos=0.1]{$\set{Q}$} (cs.189);

  \node [rectangle, right of=opt, shift={(-0.0, 0.0)}, fill opacity=0.0, text opacity=0.0, node distance=1.0cm] (k) {};
  \draw [->] (opt.east) -- node[above, pos=0.9]{$\theta^*$} (k);

  \node [rectangle, below of=cs, shift = {(-1.3, 0.5)}, node distance=1cm, fill opacity=0.0, text opacity=0.0] (k) {};
  \draw [->] (k.east) -| node[pos=0.05, left, shift={(0.0, -0.00)}]{$\theta_0$} (cs.230);

  \draw [->] (opt.south) |- ++(0.0, -0.1) -| node[below, shift={(0.05, 0.05)}]{$\theta_k$} (cs.south);

  \begin{pgfonlayer}{background}
    \path (cs.west |- cs.north)+(-0.2,0.4) node (a) {};
    \path (opt.south -| opt.east)+(+0.1,-0.4) node (b) {};
    \path[fill=color_blue!2,rounded corners, draw=color_blue, densely dotted]
      (a) rectangle (b);
  \end{pgfonlayer}
  \node [rectangle, below of=opt, node distance=1.3em, shift={(-1.85,-0.15)}] (pipeline) {\textbf{Expensive iterative estimation pipeline with residual sampling}};

  \node [block, below of=cs, shift={(-0.1, 0.4)}, draw=color_red, fill=color_red!10] (sampling) {Point\\sampling};
  \node [block, right of=sampling, shift={(0.0, -0.4)}] (cs) {Correspondence\\search};
  \draw [->] (sampling.east) -| node[above, pos=0.5, shift={(0, -0.05)}]{$\bar{\set{P}} \subseteq \set{P}$} (cs.north);

  \node [block, right of=cs, shift={(1.0, 0.0)}] (res_lin) {Residual computation\\\& linearization};
  \draw [->] (cs.east) -| ++(0.1, 0.2) -- node[above, shift={(0.0, -0.05)}]{$\mathbb{C}_{\set{Q}}^{\bar{\set{P}}} = \{ \vect{p}_i, \vect{q}_i \}$} ++(0.65, 0) |- (res_lin.west);

  \node [block, right of=res_lin, shift={(0.65, 0.0)}] (opt) {Iterative\\optimization};
  \draw [->] (res_lin.east) -| ++(0.1, 0.25) -- node[above, shift={(-0.0, -0.05)}]{$\mathbb{C}_{\set{Q}}^{\bar{\set{P}}} = \{ \vect{p}_i, \vect{q}_i, \vect{r}_i, \mat{J}_i \}$} ++(0.45, 0) |- (opt.west);

  \node[left of=sampling, shift={(0.3, 0.2)}] (label) {\scriptsize \textbf{b)}};

  \node [rectangle, left of=sampling, shift = {(-0.0, 0.0)}, node distance=0.8cm, fill opacity=0.0, text opacity=0.0] (k) {};
  \draw [->] (k.east) -- node[left, pos=0.2]{$\set{P}$} (sampling.west);

  \node [rectangle, left of=sampling, shift = {(1.5, -0.5)}, node distance=1.3cm, fill opacity=0.0, text opacity=0.0] (k) {};
  \draw [->] (k.east) -- node[left, pos=0.1]{$\set{Q}$} (cs.189);

  \node [rectangle, right of=opt, shift={(-0.0, 0.0)}, fill opacity=0.0, text opacity=0.0, node distance=1.0cm] (k) {};
  \draw [->] (opt.east) -- node[above, pos=0.9]{$\theta^*$} (k);

  \node [rectangle, below of=cs, shift = {(-1.3, 0.5)}, node distance=1cm, fill opacity=0.0, text opacity=0.0] (k) {};
  \draw [->] (k.east) -| node[pos=0.05, left, shift={(0.0, -0.00)}]{$\theta_0$} (cs.230);

  \draw [->] (opt.south) |- ++(0.0, -0.1) -| node[below, shift={(0.05, 0.05)}]{$\theta_k$} (cs.south);

  \begin{pgfonlayer}{background}
    \path (cs.west |- cs.north)+(-0.2,0.4) node (a) {};
    \path (opt.south -| opt.east)+(+0.1,-0.4) node (b) {};
    \path[fill=color_blue!2,rounded corners, draw=color_blue, densely dotted]
      (a) rectangle (b);
  \end{pgfonlayer}
  \node [rectangle, below of=opt, node distance=1.3em, shift={(-1.4,-0.15)}] (pipeline) {\textbf{Iterative estimation pipeline of reduced complexity}};

\end{tikzpicture}

%% file: fig/redundancy.tex
\tikzset{
  double -latex/.style args={#1 colored by #2 and #3}{
    -stealth,line width=#1,#2, 
    postaction={draw,-stealth,#3,line width=(#1)/3,
                shorten <=(#1)/3,shorten >=2*(#1)/3}, 
  }
}

\pgfdeclarelayer{background}
\pgfdeclarelayer{foreground}
\pgfsetlayers{background,main,foreground}

\begin{tikzpicture}[
  font=\tiny
  ]

  \node[anchor=south west,inner sep=0] (a) at (0,0) {
      \def\arraystretch{0}%
      \includegraphics[width=0.99\columnwidth]{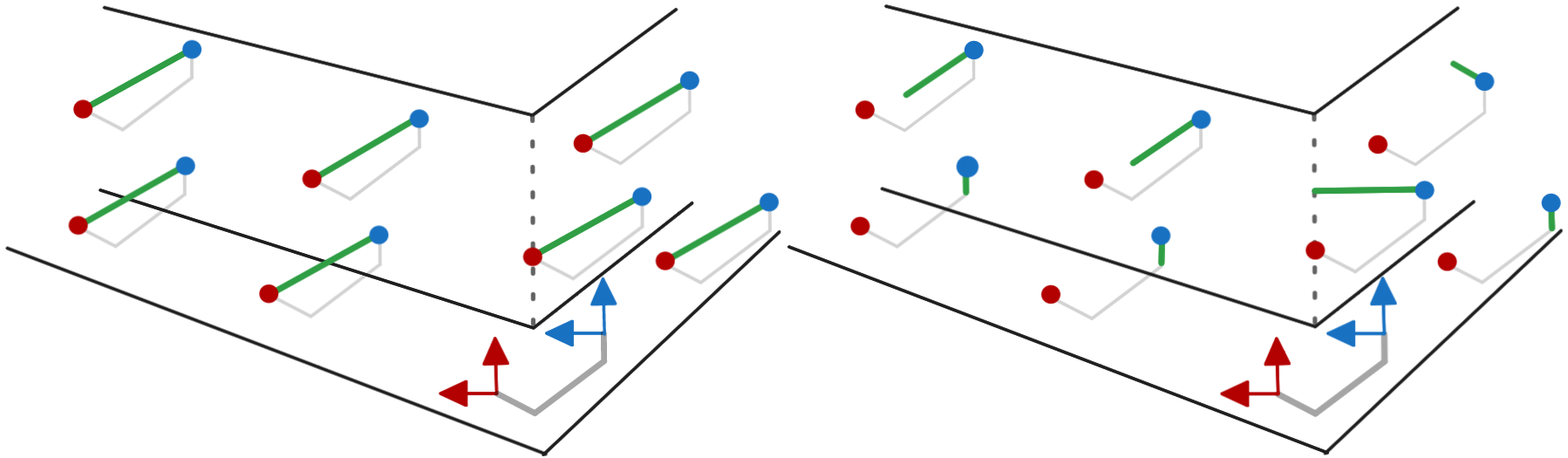}%
    }; 
  \begin{scope}[x={(a.south east)},y={(a.north west)}]


  \node[] at (0.01, 0.96) {\scriptsize \textbf{a)}};

  \node[] at (0.285, 0.21) {$\vect{t}_{k-1}$};
  \node[] at (0.400, 0.29) {$\vect{t}_{k}$};
  \node[gray] at (0.320, 0.10) {$x$};
  \node[gray] at (0.373, 0.130) {$y$};
  \node[gray] at (0.375, 0.237) {$z$};

  \node[] at (0.09, 0.87) {$\vect{r}^{\bullet}$};
  \node[] at (0.085, 0.61) {$\vect{r}^{\bullet}$};
  \node[] at (0.225, 0.72) {$\leftscript{*}{\vect{r}^{\bullet}}$};
  \node[] at (0.21, 0.47) {$\vect{r}^{\bullet}$};
  \node[] at (0.38, 0.55) {$\vect{r}^{\bullet}$};
  \node[] at (0.405, 0.79) {$\vect{r}^{\bullet}$};
  \node[] at (0.455, 0.53) {$\vect{r}^{\bullet}$};


  \node[] at (0.51, 0.96) {\scriptsize \textbf{b)}};
  
  \def\xoff{0.498}

  \node[] at (0.285+\xoff, 0.21) {$\vect{t}_{k-1}$};
  \node[] at (0.400+\xoff, 0.29) {$\vect{t}_{k}$};
  \node[gray] at (0.320+\xoff, 0.10) {$x$};
  \node[gray] at (0.373+\xoff, 0.130) {$y$};
  \node[gray] at (0.375+\xoff, 0.237) {$z$};

  \node[] at (0.085+\xoff, 0.878) {$\leftscript{*}{\vect{r}^{\square}_1}$};
  \node[] at (0.15+\xoff, 0.61) {$\leftscript{*}{\vect{r}^{\square}_2}$};
  \node[] at (0.23+\xoff, 0.73) {$\vect{r}^{\square}_1$};
  \node[] at (0.265+\xoff, 0.46) {$\vect{r}^{\square}_2$};
  \node[] at (0.325+\xoff, 0.6) {$\leftscript{*}{\vect{r}_{1}^{|}}$};
  \node[] at (0.415+\xoff, 0.78) {$\leftscript{*}{\vect{r}^{\square}_3}$};
  \node[] at (0.47+\xoff, 0.53) {$\vect{r}^{\square}_2$};

  \draw[->] (0.67, 0.82) -- node[right, shift={(-0.008, -0.03)}]{$\vect{n}_1$} (0.65, 0.755);
  \draw[->] (0.605, 0.38) -- node[right, shift={(-0.005, -0.03)}]{$\vect{n}_2$} (0.605, 0.47);
  \draw[->] (0.925, 0.94) -- node[right, shift={(-0.0, -0.0)}]{$\vect{n}_3$} (0.95, 0.89);

  \draw[->] (0.837, 0.75) -- node[right, shift={(-0.013, 0.04)}]{$\vect{v}_1$} (0.837, 0.84);


  \end{scope}

\end{tikzpicture}

%% file: fig/gfh/gfh.tex

\pgfdeclarelayer{background}
\pgfdeclarelayer{foreground}
\pgfsetlayers{background,main,foreground}

\begin{tikzpicture}[font=\tiny]

  \node[anchor=south west,inner sep=0] (a) at (0,0) {
      \def\arraystretch{0}%
      \includegraphics[width=0.98\columnwidth]{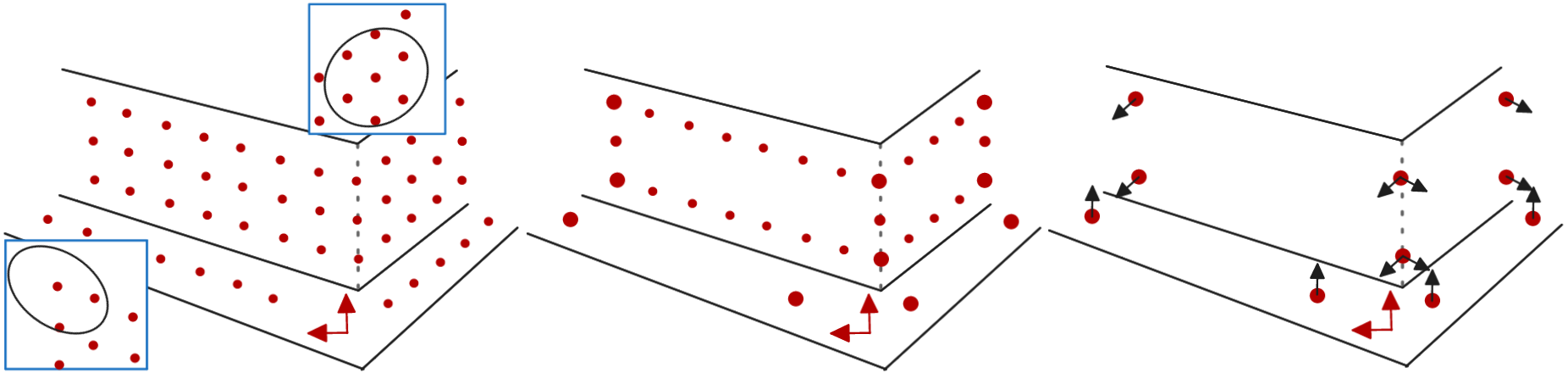}%
    }; 
  \begin{scope}[x={(a.south east)},y={(a.north west)}]


  \node[] at (0.01, 0.9) {\scriptsize \textbf{a)}};

  \draw[->, RoyalBlue] (0.04, 0.355) -- (0.059, 0.723);

  \node[] at (0.030, 0.260) {\textbf{p}};
  \node[] at (0.078, 0.220) {$\mathbf{j}_1$};
  \node[] at (0.029, 0.095) {$\mathbf{j}_2$};
    \node[] at (0.08, -0.07) {$\Delta_\vect{p} = \frac{1}{2}\hlr{\left( \vect{j}_1 + \vect{j}_2 - 2\vect{p} \right)}$};

  \draw[->, RoyalBlue] (0.2789, 0.638) -- (0.2789, 0.573);

  \node[] at (0.233, 0.8495) {$\mathbf{j}_i$};
    \node[] at (0.20, 1.06) {$\Delta_\vect{p} = \frac{1}{6}\sum_{i=1}^{6} \hlr{\left( \vect{j}_i - \vect{p} \right)} = \mathbf{0}$};


  \node[] at (0.35, 0.9) {\scriptsize \textbf{b)}};
  \node[] at (0.51, 0.9) {$\hat{\set{P}} = \left\{ \vect{p}\,|\,\Delta_{\vect{p}} > 0, \vect{p} \in \set{P}_\nu \right\}$};

  \node[] at (0.68, 0.9) {\scriptsize \textbf{c)}};
  \node[] at (0.83, 0.9) {$\hat{\set{P}} = \left\{ \vect{p}\,|\,\Delta_{\vect{p}} > \epsilon_\Delta, \vect{p} \in \set{P}_\nu \right\}$};

  \end{scope}

\end{tikzpicture}

%% file: fig/convergence_analysis/traj.tex
\pgfdeclarelayer{background}
\pgfdeclarelayer{foreground}
\pgfsetlayers{background,main,foreground}
\usetikzlibrary{arrows.meta, decorations.markings}

\begin{tikzpicture}[font=\tiny]

  \node[anchor=south west,inner sep=0] (a) at (0,0) {
      \def\arraystretch{0}%
      \includegraphics[width=1.0\textwidth,height=2.4cm]{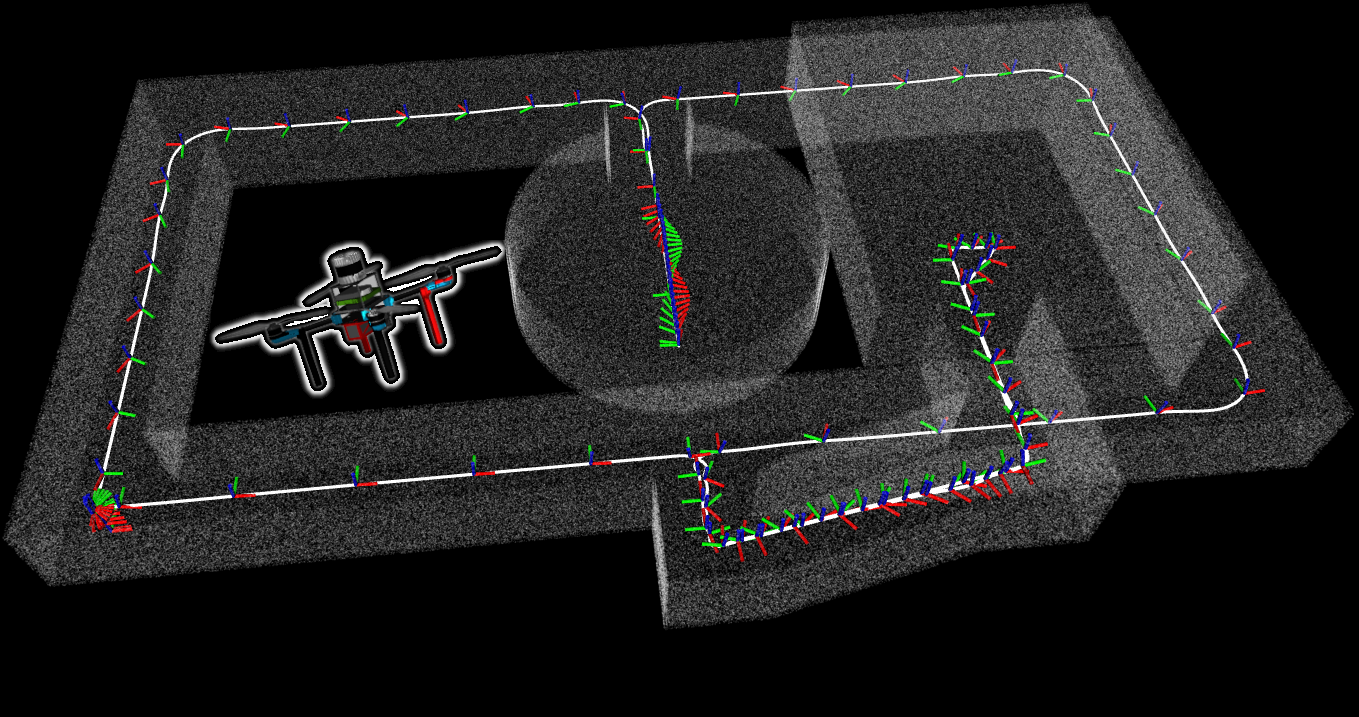}%
  }; 
  \begin{scope}[x={(a.south east)},y={(a.north west)}]



  \node[white, anchor=west] (a) at (0.148, 0.70) {\acs{mav} (15:1)};
  \node[white, anchor=west] (a) at (-0.01, 0.21) {start \& goal};

  \draw[draw=white] (0.03, 0.02) -- (0.97, 0.2);
  \node[white] (a) at (0.95, 0.125) {\SI{50}{\meter}};

  \draw[>=latex, <->, color_orange] (0.26,0.828) -- (0.36,0.842) node[midway, below, shift={(0.00, 0.02)}] {F};
  \draw[>=latex, <->, color_orange] (0.2,0.315) -- (0.3,0.33) node[midway, below, shift={(0.00, 0.02)}] {A};
  
  \draw[-latex, color_green] (0.54,0.27) -- (0.54,0.40) node[midway, right, shift={(0.0, 0)}] {B};
  \draw[
    postaction={
      decorate,
    },
    color_green
  ] (0.54,0.31) circle (0.02);

  \draw[-latex, color_orange] (0.67,0.31) -- (0.67,0.44) node[midway, right, shift={(-0.02, 0.05)}] {C};

  \draw[-latex, color_green] (0.91,0.44) -- (0.91,0.57) node[midway, right, shift={(0.00, 0)}] {D};
  \draw[
    postaction={
      decorate,
    },
    color_green
  ] (0.91, 0.48) circle (0.02);

  \draw[-latex, color_green] (0.49,0.62) -- (0.49,0.75) node[midway, right, shift={(0.00, 0)}] {E};
  \draw[
    postaction={
      decorate,
    },
    color_green
  ] (0.49, 0.66) circle (0.02);

  \draw[-latex, violet] (0.14,0.79) -- (0.14,0.92) node[midway, right, shift={(0.00, 0)}] {G};
  \draw[
    postaction={
      decorate,
    },
    violet
  ] (0.14, 0.83) circle (0.02);








  \end{scope}

\end{tikzpicture}
\vspace{-0.2cm}

%% file: fig/convergence_analysis/runtime.tex
\begin{tikzpicture}[
  font=\tiny
  ]


  \pgfplotstableread[col sep=space]{./data/convergence_analysis/011_sim_x_icp_1/runtime_data_with_manually_inserted_gfh.csv}{\tableRUNTIME}

  \begin{axis}[ 
    name=avg,
    ybar stacked,
    width=0.59\textwidth,
    height=2.7cm,
    grid=major,
    grid style={draw=gray!10,line width=.1pt},
    ylabel style={align=center},
    ylabel= average runtime (ms),
    ylabel shift=-3pt,
    y label style={xshift=-8pt},
    xtick={0, 1, 2, 3},
    xticklabels={\phantom{00000}V40, V40+NS0.2, V40+CovS80, V40+RMS},
    xticklabel style={xshift=1mm, yshift=0mm, rotate=60, anchor=east},
    ymin=0,
    enlarge x limits=0.2,
    bar width=4pt,
    ]

    \addplot [color=color_orange, fill=color_orange, bar shift=-0pt] table [y=avg_vox, x expr=\coordindex] {\tableRUNTIME};
    \addplot [color=color_green, fill=color_green, bar shift=-0pt]   table [y=avg_norm_est, x expr=\coordindex] {\tableRUNTIME};
    \addplot [color=color_red, fill=color_red, bar shift=-0pt]       table [y=avg_sel, x expr=\coordindex] {\tableRUNTIME};
    \addplot [color=color_blue, fill=color_blue, bar shift=-0pt]     table [y=avg_reg, x expr=\coordindex] {\tableRUNTIME};

  \end{axis}

  \begin{axis}[
    name=max,
    at=(avg.north east), anchor=north west,
    ybar stacked,
    width=0.59\textwidth,
    height=2.7cm,
    grid=major,
    grid style={draw=gray!10,line width=.1pt},
    ylabel style={align=center},
    ylabel=maximum runtime (ms),
    ylabel shift=-4pt,
    y label style={xshift=-10pt},
    xtick={0, 1, 2, 3},
    xticklabels={\phantom{00000}V40, V40+NS0.2, V40+CovS80, V40+RMS},
    xticklabel style={xshift=1mm, yshift=0mm, rotate=60, anchor=east},
    ymin=0,
    enlarge x limits=0.2,
    bar width=4pt,
    xshift=0.9cm,
    legend columns=2,
    legend image post style={scale=0.45},
    legend style={text=black, draw=none, fill=none, at={(1.1, 1.5)}, legend cell align=left, row sep=-3pt},
    ]

    \addplot [color=color_orange, fill=color_orange, bar shift=-0pt] table [y=max_vox, x expr=\coordindex] {\tableRUNTIME};
    \addplot [color=color_green, fill=color_green, bar shift=-0pt]   table [y=max_norm_est, x expr=\coordindex] {\tableRUNTIME};
    \addplot [color=color_red, fill=color_red, bar shift=-0pt]       table [y=max_sel, x expr=\coordindex] {\tableRUNTIME};
    \addplot [color=color_blue, fill=color_blue, bar shift=-0pt]     table [y=max_reg, x expr=\coordindex] {\tableRUNTIME};

    \addlegendentry{voxelization}
    \addlegendentry{\hlr{\acs{gfh}\,|\,normals extraction}}
    \addlegendentry{sampling}
    \addlegendentry{\hlr{optimization}}

  \end{axis}

\end{tikzpicture}

%% file: fig/convergence_analysis/performance.tex
\pgfplotsset{compat=newest, 
  emphasize/.code args={#1:#2with#3}{
    \pgfplotsextra{
            \draw[color=#3, fill=#3, fill opacity=0.15, draw opacity=0.0] ({axis cs:#1,0.00} |- {axis description cs:0,0.006}) 
            rectangle ({axis cs:#2,0} |- {axis description cs:0,0.994});
    }
  }
}

\begin{tikzpicture}[
    >=latex,
    font=\tiny
  ]

  \pgfplotstableread[col sep=space]{./data/convergence_analysis/011_sim_x_icp_1/VOXELIZATION_evo.txt}{\tableVANILLAEVO}
  \pgfplotstableread[col sep=space]{./data/convergence_analysis/011_sim_x_icp_1/RISE_evo.txt}{\tableRISDEVO}
  \pgfplotstableread[col sep=space]{./data/convergence_analysis/011_sim_x_icp_1/NORMAL_SAMPLING_evo.txt}{\tableNSEVO}
  \pgfplotstableread[col sep=space]{./data/convergence_analysis/011_sim_x_icp_1/COVARIANCE_SAMPLING_evo.txt}{\tableCOVSEVO}

  \pgfplotstableread[col sep=space]{./data/convergence_analysis/011_sim_x_icp_1/VOXELIZATION.csv}{\tableVANILLA}
  \pgfplotstableread[col sep=space]{./data/convergence_analysis/011_sim_x_icp_1/RISE.csv}{\tableRISD}
  \pgfplotstableread[col sep=space]{./data/convergence_analysis/011_sim_x_icp_1/NORMAL_SAMPLING.csv}{\tableNS}
  \pgfplotstableread[col sep=space]{./data/convergence_analysis/011_sim_x_icp_1/COVARIANCE_SAMPLING.csv}{\tableCOVS}

  \pgfplotsset{
    legend image code/.code={
      \draw[]
      plot coordinates {
        (0cm,0cm)
        (0.15cm,0cm)        
        (0.3cm,0cm)         
      };%
    }
  }

  \begin{axis}[ 
    name=err,
    width=0.57\textwidth,
    height=2.7cm,
    grid=major,
    grid style={draw=gray!30,line width=.1pt},
    ylabel=APE (m),
    ylabel shift=-3pt,
    ymax=2.5,
    xticklabels={},
    legend columns=4,
    legend style={text=black, fill=none, draw=none, at={(1.95, 1.32)}, row sep=-3pt, legend cell align=left},
    enlarge x limits=0,
    axis line style={latex-latex},
    ]
    \addplot [smooth, color=color_blue, line width=0.7pt,
    emphasize=10:20 with color_orange, 
    emphasize=21:26.5 with color_green, 
    emphasize=56:62 with color_green, 
    emphasize=29.5:37 with color_orange, 
    emphasize=48:55.5 with color_orange, 
    emphasize=75.5:79.5 with color_green, 
    emphasize=91:113 with color_green, 
    emphasize=124:128 with violet, 
    emphasize=114:122 with color_orange, 
    ] table [y=ape, x=sec_from_start]{\tableVANILLAEVO};
    \addplot [smooth, color=color_green, line width=0.7pt] table [y=ape, x=sec_from_start]{\tableNSEVO};
    \addplot [smooth, color=black, line width=0.7pt] table [y=ape, x=sec_from_start]{\tableCOVSEVO};
    \addplot [smooth, color=color_red, line width=0.7pt] table [y=ape, x=sec_from_start]{\tableRISDEVO};

    \addlegendentry{\tiny{V40}}
    \addlegendentry{\tiny{V40+NS0.2}}
    \addlegendentry{\tiny{V40+CovS80}}
    \addlegendentry{\tiny{V40+RMS (proposed)}}

    \node [draw, fill=white, fill opacity=0.8, draw opacity=1.0, text opacity=1.0, align=left, anchor=north west, inner sep=1pt] at (0, 2.5) {
      \parbox{0.9cm}{mean (\si{\meter}):} \textcolor{color_blue}{0.27} \textcolor{color_green}{0.44} \textcolor{black}{0.33} \textcolor{color_red}{\textbf{0.19}} \\
      \parbox{0.9cm}{RMSE (\si{\meter}):} \textcolor{color_blue}{0.38} \textcolor{color_green}{0.52} \textcolor{black}{0.44} \textcolor{color_red}{\textbf{0.24}} \\
      \parbox{0.9cm}{max (\si{\meter}):}  \textcolor{color_blue}{1.07} \textcolor{color_green}{1.32} \textcolor{black}{1.15} \textcolor{color_red}{\textbf{0.61}}
    };

  \end{axis}

  \begin{axis}[ 
    name=timing,
    at=(err.north east), anchor=north west,
    width=0.57\textwidth,
    height=2.7cm,
    grid=major,
    grid style={draw=gray!30,line width=.1pt},
    xlabel shift=-7pt,
    ylabel style={align=center},
    ylabel=total\\runtime (\si{\milli\second}),
    ylabel shift=-5pt,
    xticklabels={},
    enlarge x limits=0,
    axis line style={latex-latex},
    xshift=1.0cm,
    ymin=0,
    ]
    \addplot [smooth, color=color_blue, line width=0.7pt, draw opacity=1.0,
    emphasize=10:20 with color_orange, 
    emphasize=21:26.5 with color_green, 
    emphasize=56:62 with color_green, 
    emphasize=29.5:37 with color_orange, 
    emphasize=48:55.5 with color_orange, 
    emphasize=75.5:79.5 with color_green, 
    emphasize=91:113 with color_green, 
    emphasize=114:122 with color_orange, 
    emphasize=124:128 with violet, 
    ] table [y=t, x=ts]{\tableVANILLA};
    \addplot [smooth, color=color_green, line width=0.7pt, draw opacity=1.0] table [y=t, x=ts]{\tableNS};
    \addplot [smooth, color=black, line width=0.7pt, draw opacity=1.0] table [y=t, x=ts]{\tableCOVS};
    \addplot [smooth, color=color_red, line width=0.7pt, draw opacity=1.0] table [y=t, x=ts]{\tableRISD};

    \node[black] at (15, 110) {\textbf{A}};
    \node[black] at (23.75, 110) {\textbf{B}};
    \node[black] at (59, 110) {\textbf{B}};
    \node[black] at (33.25, 110) {\textbf{C}};
    \node[black] at (51.75, 110) {\textbf{C}};
    \node[black] at (77.5, 110) {\textbf{D}};
    \node[black] at (109, 144) {\textbf{E}};
    \node[black] at (126, 110) {\textbf{G}};
    \node[black] at (118, 110) {\textbf{F}};

    \node [draw, fill=white, fill opacity=0.8, draw opacity=1.0, text opacity=1.0, align=left, anchor=north west, inner sep=1pt] at (0, 160) {
      mean (\si{\milli\second}): \textcolor{color_blue}{37.9} \textcolor{color_green}{36.1} \textcolor{black}{39.3} \textcolor{color_red}{\textbf{29.3}}
    };

  \end{axis}

  \begin{axis}[ 
    name=compr,
    at=(err.below south west), anchor=above north west,
    width=0.57\textwidth,
    height=2.4cm,
    grid=major,
    grid style={draw=gray!30,line width=.1pt},
    xlabel=time (s),
    xlabel shift=-7pt,
    ylabel style={align=center},
    ylabel=compr.\\rate (\si{\percent}),
    ylabel shift=-5pt,
    enlarge x limits=0,
    axis line style={latex-latex},
    ymin=91,
    yshift=0.55em,
    ]
    \addplot [smooth, color=color_blue, line width=0.7pt,
    emphasize=10:20 with color_orange, 
    emphasize=21:26.5 with color_green, 
    emphasize=56:62 with color_green, 
    emphasize=29.5:37 with color_orange, 
    emphasize=48:55.5 with color_orange, 
    emphasize=75.5:79.5 with color_green, 
    emphasize=91:113 with color_green, 
    emphasize=124:128 with violet, 
    emphasize=114:122 with color_orange, 
    ] table [y=compr, x=ts]{\tableVANILLA};
    \addplot [smooth, color=color_green, line width=0.7pt] table [y=compr, x=ts]{\tableNS};
    \addplot [smooth, color=black, line width=0.7pt] table [y=compr, x=ts]{\tableCOVS};
    \addplot [smooth, color=color_red, line width=0.7pt] table [y=compr, x=ts]{\tableRISD};

    \node [draw, fill=white, fill opacity=0.8, draw opacity=1.0, text opacity=1.0, align=left, anchor=south west, inner sep=1pt] at (0, 91) {
      mean (\si{\percent}): \textcolor{color_blue}{96.2} \textcolor{color_green}{98.1} \textcolor{black}{96.7} \textcolor{color_red}{\textbf{98.2}}
    };

  \end{axis}

  \begin{axis}[ 
    name=ez,
    at=(compr.north east), anchor=north west,
    width=0.57\textwidth,
    height=2.4cm,
    grid=major,
    grid style={draw=gray!30,line width=.1pt},
    xlabel=time (s),
    xlabel shift=-7pt,
    ylabel=$R_z$ (-),
    ylabel shift=-3pt,
    ytick={0, 150},
    enlarge x limits=0,
    enlarge y limits=0.2,
    axis line style={latex-latex},
    xshift=1.0cm,
    ]
    \addplot [smooth, color=color_blue, line width=0.7pt,
    emphasize=10:20 with color_orange, 
    emphasize=21:26.5 with color_green, 
    emphasize=56:62 with color_green, 
    emphasize=29.5:37 with color_orange, 
    emphasize=48:55.5 with color_orange, 
    emphasize=75.5:79.5 with color_green, %
    emphasize=91:113 with color_green, 
    emphasize=124:128 with violet, 
    emphasize=114:122 with color_orange, 
    ] table [y=e_z, x=ts]{\tableVANILLA};
    \addplot [smooth, color=color_green, line width=0.7pt] table [y=e_z, x=ts]{\tableNS};
    \addplot [smooth, color=black, line width=0.7pt] table [y=e_z, x=ts]{\tableCOVS};
    \addplot [smooth, color=color_red, line width=0.7pt, draw opacity=0.8] table [y=e_z, x=ts]{\tableRISD};
    
    \node [draw, fill=white, fill opacity=0.8, draw opacity=1.0, text opacity=1.0, align=left, anchor=north west, inner sep=1pt] at (0, 292) {
      rel. mean (\si{\percent}): \textcolor{color_blue}{0.94} \textcolor{color_green}{0.93} \textcolor{black}{0.90} \textcolor{color_red}{\textbf{1.0}}
    };

  \end{axis}

\end{tikzpicture}

%% file: fig/gfh/gfh_analysis.tex
\begin{tikzpicture}[
    >=latex,
    font=\scriptsize
  ]

  \pgfplotstableread[col sep=space]{./data/gfh/data_600.csv}{\table}
  \pgfplotstableread[col sep=space]{./data/gfh/data_600_sel.csv}{\tableSEL}
  
  \begin{axis}[ 
    name=ax1,
    width=1.0\columnwidth,
    height=0.27\columnwidth,
    grid=major,
    grid style={draw=gray!30,line width=.1pt},
    ylabel style={align=center},
    ylabel=\hlg{(\si{\percent})},
    xmin=5,
    ytick={0, 0.5, 1},
    yticklabels={0, 50, 100},
    ylabel shift=-5pt,
    xticklabels={},
    legend columns=3,
    legend style={text=black, fill=none, draw=none, at={(1.0, 0.84)}},
    enlarge x limits=0,
    enlarge y limits=0.2,
    axis line style={latex-latex},
    xmode=log,
    x tick style={color=gray!30},
    y tick style={color=gray!30},
    ]
    \draw[thick] (10, -0.2) -- (10, 1.2) node [midway, left, xshift=0.07cm] (TextNode) {\tiny{K}};
    \draw[thick] (100000, 0.004) -- (0, 0.004) node [above, xshift=0.65cm, yshift=-0.08cm] (TextNode) {\tiny{$\lambda_H$}};

    \addplot [smooth, color=color_blue, line width=0.7pt, densely dashed, forget plot] table [y=entropy, x=count]{\table};
    \addplot [smooth, color=color_red, line width=0.7pt, densely dashed, forget plot] table [y=rate_rel, x=count]{\table};

    \addplot [smooth, color=color_blue, line width=1.0pt] table [y=entropy, x=count]{\tableSEL};
    \addplot [smooth, color=color_red, line width=1.0pt] table [y=rate_rel, x=count]{\tableSEL};
    
    \addlegendentry{\tiny{\hlg{relative entropy $\frac{H_\Delta}{\text{max}(H_\Delta)}$}}}
    \addlegendentry{\tiny{relative rate $\bar{H}_\Delta$}}

  \end{axis}

  \begin{axis}[ 
    name=ax2,
    at=(ax1.below south west), anchor=above north west,
    width=1.0\columnwidth,
    height=0.23\columnwidth,
    grid=major,
    grid style={draw=gray!30,line width=.1pt},
    x tick style={color=gray!30},
    y tick style={color=gray!30},
    xlabel=sampled point count (-),
    xlabel shift=-5pt,
    ylabel style={align=center},
    ylabel=\hlg{(\si{\percent})},
    ylabel shift=-5pt,
    xmin=5,
    ytick={0.9, 1},
    yticklabels={90, 100},
    enlarge x limits=0,
    enlarge y limits=0.2,
    axis line style={latex-latex},
    legend style={text=black, fill=none, draw=none, at={(1.00, 1.05)}},
    yshift=1.0em,
    xmode=log,
    ]
    \draw[thick] (10, 0.8) -- (10, 1.2) node [midway, left, xshift=0.07cm, yshift=-1.0mm] (TextNode) {\tiny{K}};

    \addplot [smooth, color=color_green, line width=0.7pt, densely dashed, forget plot] table [y=red_rel, x=count]{\table};
    \addplot [smooth, color=color_green, line width=1.0pt] table [y=red_rel, x=count]{\tableSEL};

    \addlegendentry{\tiny{relative redundancy \hlg{$\frac{R - \bar{\mu}_\Delta}{R} \left( R = \log_{10}K \right)$}}}
  \end{axis}

\end{tikzpicture}